%% file: main.tex
\documentclass[sigconf]{acmart}

\AtBeginDocument{%
  \providecommand\BibTeX{{%
    \normalfont B\kern-0.5em{\scshape i\kern-0.25em b}\kern-0.8em\TeX}}}

\usepackage[ruled,linesnumbered]{algorithm2e}
\usepackage{algpseudocode}
\usepackage{bbding}
\usepackage{CJKutf8}
\usepackage{enumitem}
\usepackage{tabularray}
\usepackage{makecell}
\usepackage{graphicx}
\usepackage{subfigure}
\usepackage{latexsym}
\usepackage{xcolor}
\usepackage{caption}
\usepackage{booktabs}
\usepackage{multirow}
\usepackage[normalem]{ulem}
\usepackage{hyperref}
\usepackage{float}
\input{math_commands.tex}

\usepackage{newfloat}
\usepackage{listings}
\usepackage{svg}
\usepackage{stfloats}

\usepackage{threeparttable}
\usepackage{colortbl}

\definecolor{mygray}{gray}{0.85}

\usepackage{xspace}

\newcommand{\vpara}[1]{\vspace{0.04in}\noindent\textbf{#1}\xspace}

\newcommand{\eg}{e.g.}
\newcommand{\model}{HGVAE\xspace}

\usepackage{colortbl}
\definecolor{mygrey}{gray}{0.925}

\begin{document}

\title{\model: Integrating Generative and Contrastive Learning for Heterogeneous Graph Learning}
\title{Refining Latent Representations: A Generative SSL Approach for Heterogeneous Graph Learning}

\author{Yulan Hu}
\email{huyulan@ruc.edu.cn}
\affiliation{%
  \institution{Gaoling School of Artificial Intelligence,  Renmin University of China}  
  \city{Beijing}  
  \country{China}  
}

\author{Zhirui Yang}
\email{yangzhirui@ruc.edu.cn}
\affiliation{%
  \institution{Gaoling School of Artificial Intelligence,  Renmin University of China}  
  \city{Beijing}  
  \country{China}  
}
 
\author{Sheng Ouyang}
\email{ouyangsheng@ruc.edu.cn} 
\affiliation{%
  \institution{Gaoling School of Artificial Intelligence,  Renmin University of China}  
  \city{Beijing}  
  \country{China}  
}

\author{Yong Liu}
\email{liuyonggsai@ruc.edu.cn}
\affiliation{%
  \institution{Gaoling School of Artificial Intelligence,  Renmin University of China}  
  \city{Beijing}  
  \country{China}  
}

\renewcommand{\shortauthors}{Yulan Hu et al.}


\begin{abstract}
Self-Supervised Learning (SSL) has shown significant potential and has garnered increasing interest in graph learning. However, particularly for generative SSL methods, its potential in Heterogeneous Graph Learning (HGL) remains relatively underexplored. Generative SSL utilizes an encoder to map the input graph into a latent representation and a decoder to recover the input graph from the latent representation. Previous HGL SSL methods generally design complex strategies to capture graph heterogeneity, which heavily rely on contrastive view construction strategies that are often non-trivial. Yet, refining the latent representation in generative SSL can effectively improve graph learning results. In this study, we propose \model, a generative SSL method specially designed for HGL. Instead of focusing on designing complex strategies to capture heterogeneity, \model centers on refining the latent representation. Specifically, \model innovatively develops a contrastive task based on the latent representation. To ensure the hardness of negative samples, we develop a progressive negative sample generation (PNSG) mechanism that leverages the ability of Variational Inference (VI) to generate high-quality negative samples. As a pioneer in applying generative SSL for HGL, \model refines the latent representation, thereby compelling the model to learn high-quality representations. Compared with various state-of-the-art (SOTA) baselines, \model achieves impressive results, thus validating its superiority. 

\end{abstract}



\maketitle

\section{Introduction}\label{sec:intro}
Heterogeneous information networks (HINs)~\cite{hin} are prevalent in the real world\cite{cai2005mining, jin2021application, shi2016survey}, which encompass various node and edge types. To tackle the challenge of heterogeneity and effectively capturing the valuable semantic knowledge within HINs, several heterogeneous information neural networks (HGNNs) have been proposed. These include both unsupervised methods, which design meta-paths~\cite{HAN, hgt, hetgnn, magnn}, and supervised methods~\cite{Simple-HGN} that rely on labeled information. In recent years, SSL has emerged as a promising approach in heterogeneous graph learning (HGL)~\cite{HeCo, hgnnac, HGMAE}. SSL methods extract information from graph data itself, eliminating the need for supervised information.

The utilization of SSL in graph learning can be broadly categorized into two groups: contrastive SSL methods~\cite{Grace, GCA, GraphCL, DGI2018, GCC, HeCo} and generative SSL methods~\cite{VGAE, GALA, GraphMAE, HGMAE, vigraph}. However, the incorporation of SSL in HGL remains relatively unexplored. With HGL, contrastive SSL methods primarily construct contrastive views that discriminate semantically similar samples from dissimilar samples (negatives) in latent space. For instance, Heco~\cite{HeCo} builds network schema views and meta-path views, subsequently performing contrastive SSL among these views. In contrast to contrastive SSL, generative SSL aims to reconstruct input data and utilize it for supervision, although its applications in HGL are relatively rare. HGMAE~\cite{HGMAE} represents a recent effort to introduce generative SSL to HGL, adopting masked graph autoencoder (GAE) as a backbone, along with comprehensive training strategies to extract heterogeneous knowledge.

Despite preliminary attempts to apply SSL methods to graph learning have shown promising results, certain limitations still persist. For contrastive SSL, existing HGL studies predominantly focus on constructing diverse meta-paths to establish cross-view contrastive objectives. For instance, considering the DBLP citation network encompassing Term (T), Paper (P), Author (A), and Conference (C) nodes, two meta-path views, namely APA and ACPCA, are created. However, these views may share similar attributes from overlapping nodes, such as Author and Paper nodes. Recent research~\cite{contrastive_hard, zhu2022structure, progcl} has revealed the benefits of incorporating hard negative samples — samples semantically close to the anchor node but which belong to distinct categories — on performance improvement. Yet, in the HGL context, no research has been undertaken on constructing high-quality hard negative samples for contrastive SSL. Consequently, the problem of how to construct effective contrastive views in HGL remains unsolved.

Furthermore, for generative SSL, it offers several advantages over contrastive SSL. One of the most notable advantages is the elimination of the need to construct contrastive views, as it captures implicit graph knowledge through reconstruction. However, the applications of generative SSL in graph learning exhibits two limitations. On one hand, compared to generative applications in Computer Vision (CV)~\cite{mae, clip, dalle2} and Natural Language Processing (NLP)~\cite{simcse, bert,  gpt}, the potential of generative SSL in graph learning remains relatively underdeveloped. In the realm of HGL, only a few works, such as HGMAE~\cite{HGMAE}, were developed, indicating the need for greater attention to this area. On the other hand, the existing generative graph SSL methods, \eg, VGAE~\cite{VGAE}, GraphMAE~\cite{GraphMAE}, and HGMAE~\cite{HGMAE}, employ autoencoders to align inputs and outputs with complex strategies during the reconstruction process. For example, HGMAE~\cite{HGMAE} develops three training strategies and two masking mechanisms to mine heterogeneous knowledge and enable robust training, while limited investigation for the refinement of latent representations have been made. The previous work, AGRA~\cite{arga}, adversarially aligns latent representations with prior Gaussian noise, enhancing the overall performance of node-level graph tasks. Despite that AGGA revealed the benefits of latent representation refinement, there have been no subsequent work that further investigate this.

Based on the preceding analysis, we put forward the question, \textbf{\textit{Can a generative SSL approach for HGL achieve promising performance by effectively exploiting the latent representation?}} This inquiry involves considerations from three distinct aspects.

\textit{First, how can we design a model that serves as the foundational framework for generative latent representation learning?} Previous generative methods~\cite{GraphMAE, s2gae, graphmae2, HGMAE} have primarily relied on Graph Autoencoders (GAEs) as the base model. However, these models, built upon autoencoders, may not fully exploit their generative potential. In Computer Vision (CV), Variational Autoencoders (VAEs)~\cite{VAE} are utilized to map input features to a hidden distribution, generating realistic images by sampling from this distribution~\cite{vaecv}. In contrast, Variational Graph Autoencoder (VGAE) employs variational inference (VI) to generate latent variables for the latent representation. The enhancement in the quality of samples generated by VI potentially improves the overall quality of the graph representation. Therefore, we opt for VGAE as the base model to fully leverage its generative capabilities.

\textit{Second, how can we effectively utilize latent representations?} The existing approaches have predominantly focused on designing strategies to characterize graph features. In fact, based on the learned latent variables, we can generate multiple latent representations, which can be used to establish cross-view contrastive learning tasks, thereby improving the quality of the latent representation. Guided by this principle, we devise an efficient learning strategy that harnesses the capabilities of Variational Inference (VI), allowing us to employ two forms of SSL tasks—generative and contrastive—to refine the overall latent representation. Specifically, the contrastive task involves constructing triples consisting of anchor, positive, and negative samples. Drawing inspiration from SimCSE~\cite{simcse}, we generate anchor and positive samples through two-time encoding with differing hyperparameters, resulting in two encoded views that resemble each other but are not identical. In addition, the quality of negative samples—the number and hardness of negative samples—strongly influences contrastive SSL~\cite{hardnegative, progcl}. This emphasis promotes the formation of a difficult contrastive task, which is beneficial for overall model learning.

\textit{Third, how to construct a difficult contrastive graph SSL task?} The difficulty of contrastive learning primarily lies in the hardness of the negative samples~\cite{hardnegative, progcl}. High-quality hard negative samples should be very close to the anchor node but belong to a different category, compelling the model to learn a robust representation. Existing graph SSL methods typically construct negative samples based on existing nodes in the graph~\cite{Grace, GCA, HeCo}, or synthesize new nodes using mixture methods between adjacent similar nodes~\cite{progcl}. However, these methods often introduce false negative samples that hinder model optimization. Conversely, we propose a novel adaptive method, Progressive Negative Samples Generation (PNSG), for generating hard negative samples. PNSG produces negative samples at different training stages in a progressive manner, incorporating samples generated by VI from a shifted stochastic variable as well as randomly corrupted samples from the encoded representation. The proportion of negative samples from these two parts dynamically changes during training. At early training stages, when the model is not sufficiently trained, the quality of negative samples generated by VI is relatively poor. Therefore, corrupted samples account for a larger proportion at this stage. As training progresses and the quality of generated samples improves, we increase the proportion of samples generated by VI. This dynamic adaptive adjustment ensures the model consistently encompasses hard negative samples. Consequently, as the training progresses and the quality of the generated negative samples improves, the model is compelled to adjust to increasingly stringent conditions.
  
Our proposed model, \model, effectively addresses the aforementioned three questions. Our contributions are summarized as:
\begin{itemize}
\item \textbf{Generative SSL in HGL.} We pioneer the application of generative SSL in HGL, and for the first time, we introduce contrastive SSL into generative SSL, collaboratively enhancing HGL. Building upon this, we propose \model for HGL.
\item \textbf{Progressive negative samples generation.} We develop a progressive negative samples generation strategy, which innovatively leverages the generative ability of VI to ensure the hardness of the negative samples.
\item \textbf{Experimental validation.} We conduct extensive experiments on various HGL tasks across multiple datasets. \model achieves promising performance against other SOTA HGL approaches, demonstrates its effectiveness.
\end{itemize}

\section{RELATED WORKS}
We introduce the related works of HGL, generative SSL and contrastive SSL in graph learning in the following paragraphs.

\vpara{Heterogeneous graph learning.} Heterogeneous graph neural networks (HGNNs) have attracted considerable attention in recent years~\cite{survey1, survey2}. Random walk-based methods can be regarded as the pioneers that learn the latent representation of the graph and co-occurrence probability between nodes~\cite{line, mp2vec,node2vec}. In the last few years, methods based on GNNs and meta-paths (or a combination of both) have been proposed~\cite{HAN, DMGI, magnn, Simple-HGN, SeHGNN}. Notably, HAN~\cite{HAN} is a typical work that introduces a hierarchical attention mechanism to learn node-level and meta-path-based semantic-level features. Building upon this foundation, MAGNN~\cite{magnn} proposes to leverage intra-meta-path aggregation to incorporate intermediate semantic nodes and inter-meta-path aggregation to combine messages from multiple meta-paths. HetGNN~\cite{hetgnn} creatively uses random walks to create training samples and aggregates adjacent node attributes for feature representation. Furthermore, Simple-HGN~\cite{Simple-HGN} revisits the development of HGNNs and constructs diverse heterogeneous graph benchmarks to facilitate robust and reproducible HGNN research. SeHGNN~\cite{SeHGNN} proposed a simple but effective meta-path-based framework and achieved promising results on several benchmarks. More recently, HGMAE~\cite{HGMAE} emerged as a generation-based SSL method that utilizes a masked GAE with diverse strategies to enhance HGL.

\vpara{Contrastive SSL in graph learning.} Contrastive graph learning utilizes positive and negative sample pairs from the graph for valuable node representation, eliminating the need for labels. Most graph contrastive SSL methods are tailored for homogeneous graphs. Grace~\cite{Grace} presents a simple, potent contrastive framework that establishes two graph views by eliminating edges and corrupting node features, and maximizes the node representation consistency across these views using a contrastive loss. GCA~\cite{GCA} enhances this framework by introducing adaptive augmentation that assimilates various graph topology and semantics priors. GraphCL~\cite{GraphCL} extends SimCLR~\cite{SimCLR} concepts, from the vision domain, for graph-structured data, constructing different views through multiple graph augmentation techniques and maximizing mutual information between views. Notably, some contrastive graph learning efforts are targeting heterogeneous graphs. DMGI~\cite{DMGI}, influenced by Deep Graph Infomax (DGI)~\cite{DGI2018}, offers a systematic solution for node embedding integration across multiple graphs via the consensus regularization framework and universal discriminator. 
HeCo~\cite{HeCo} creates two graph views from the network schema and meta-path perspectives for heterogenous graph information capture. STENCIL~\cite{zhu2022structure} proposes an innovative multi-view contrastive aggregation objective for adaptive information extraction and uses structural embeddings to identify true and challenging negative samples.

\vpara{Generative SSL in graph learning.} Generative SSL methods aim to reconstruct the input given the contexts. In graph learning, VGAE~\cite{VGAE} represents the pioneering work that utilizes generative models for graph tasks. It employs a two-layer GCN~\cite{gcn} as an encoder and an inner product as a decoder for link prediction. Another approach, MGAE~\cite{mgae}, uses a marginalized GCN to learn graph embeddings, which are then optimized by minimizing the distance between the graph embeddings and input features. The works of GALA~\cite{GALA} also focus on feature reconstruction but in a symmetric autoencoder form. Later, GraphMAE~\cite{GraphMAE} developed the idea of masked auto-encoders (MAE) from computer vision to graph learning. It adopts two rounds of mask strategies to enable robust training. GraphMAE2~\cite{graphmae2} is built upon GraphMAE for robust generative SSL with multi-view reconstruction. S2GAE~\cite{s2gae} focuses on edge reconstruction as objective. For the heterogeneous domain, HGMAE~\cite{HGMAE} further extends the idea of MAE to HGNNs and demonstrates encouraging performance on several benchmarks.

\section{PRELIMINARIES}
In this section, we present the notations and problem definition, we also introduce the necessary background in Appendix~\ref{apx:background}, including the variational graph autoencoder (VGAE)~\cite{VGAE} and heterogeneous graph attention network (HAN)~\cite{HAN}.
\subsection{Notations}
For HINs, we consider a graph $\mathcal{G} = (\mathcal{V}, \mathcal{E}, \phi, \psi, \mathcal{X})$, where $\mathcal{V}$ represents the set of $N$ nodes, and $\mathcal{E} \subseteq \mathcal{V} \times \mathcal{V}$ is the set of edges connecting the nodes. Each node $v$ is associated with a type denoted by $\phi(v)$, and each edge $e$ has a type represented by $\psi(e)$. The sets of possible node types and edge types are denoted by $\mathcal{T}_{v}=\{\phi(v): \forall v \in \mathcal{V}\}$ and $\mathcal{T}_{e}=\{\psi(e): \forall e \in \mathcal{E}\}$ respectively, where $\left|\mathcal{T}_{v}\right|+\left|\mathcal{T}_{e}\right|>2$. When $\left|\mathcal{T}_{v}\right|=\left|\mathcal{T}_{e}\right|=1$, the graph degenerates into a homogeneous graph. The node features are represented by $\mathcal{X} \in \mathbb{R}^{N \times F_s}$, where $F_s$ is the dimension of the node features. Each node $v \in \mathcal{V}$ is associated with a feature vector $x_v \in \mathcal{X}$, and each edge $e_{u,v} \in \mathcal{E}$ represents a connection between node $u$ and node $v$. The detailed introduction of notations used in this paper is presented in Table~\ref{method:notation}.

\subsection{Problem Definition}
Given a heterogeneous graph $\mathcal{G}$, the task is to learn graph representations for nodes of the target type $\phi_{T}$, which will subsequently be used for downstream tasks. In this paper, we approach this problem from a generative perspective, designing an encoder $f_E$ to characterize the complex graph attributes initially, followed by a decoder $f_D$ to reconstruct the input features. Finally, we utilize $f_E$ to generate the target representations for downstream tasks.

\section{METHODOLOGY}
\begin{figure*}[h]
\centering 
\includegraphics[width=0.9\textwidth]{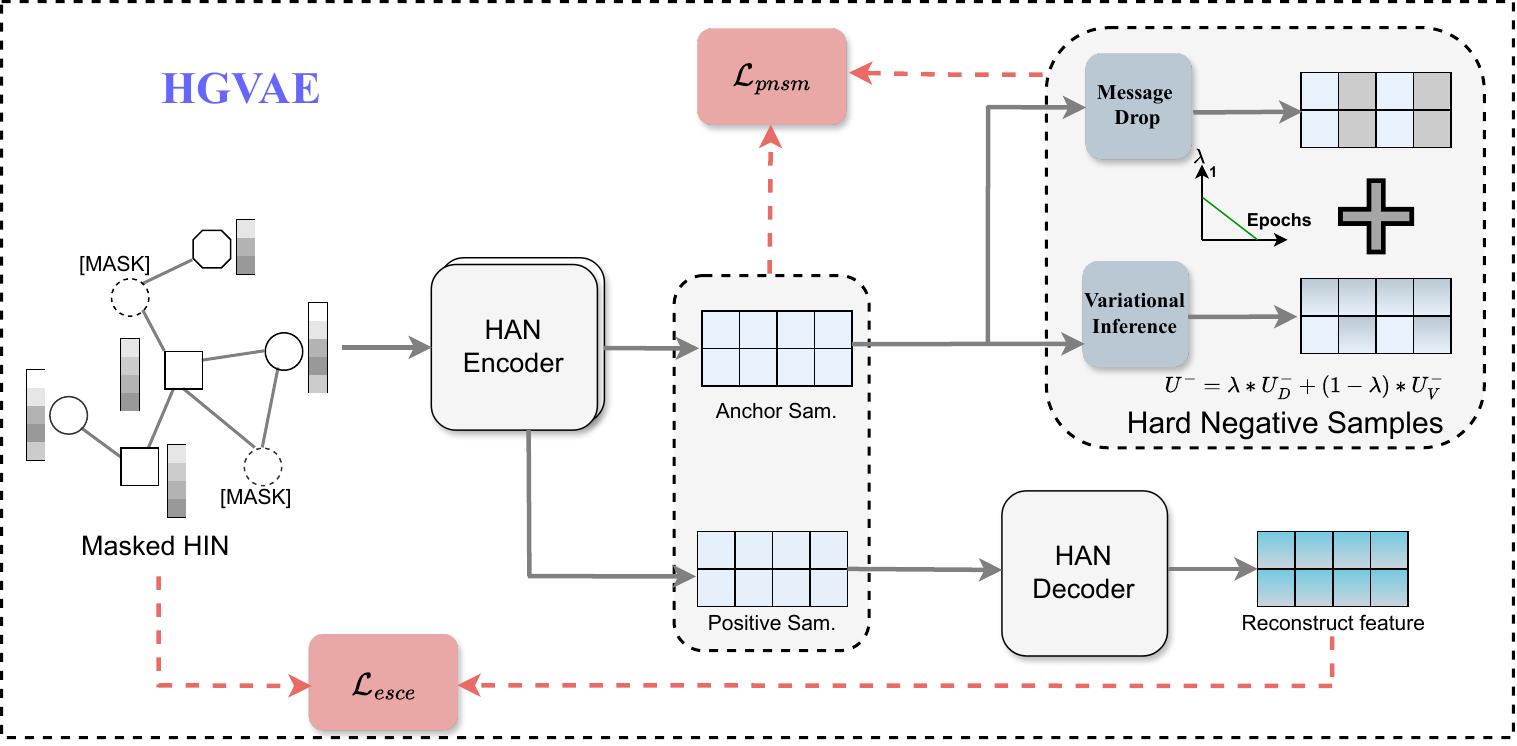} 
\caption{The Overview of \model. Given an input HIN, the HAN encoder transforms it into two distinct views: an anchor and a positive view. Subsequently, the PNSG strategy is employed to generate hard negative samples (via message dropout and Variational Inference) in a progressive manner, based on the selected view (illustrated as the anchor samples in the figure). Ultimately, contrastive SSL is conducted using the latent representation, incorporating the anchor, positive, and negative samples. Conversely, another view (depicted as the positive sample in the figure) is selected for decoding, and reconstruction is performed based on the decoding results.} 
\label{fig:model_stru}
\end{figure*}

In this section, we propose \model as a novel solution harnessing the potential of generative SSL to address the problem of HGL. The variational encoding process will be introduced in Section~\ref{method:encoding}. Subsequently, we present the latent representation contrastive learning in Section~\ref{method:contrastive_learning}. Next, we introduce the decoding process in Section~\ref{method:decoding}, followed by the illustration of training objectives in Section~\ref{method:training_objective}. The overall framework of \model is illustrated in Figure~\ref{fig:model_stru}.

\subsection{Variational Graph Autoencoder}~\label{method:encoding}
During the encoding process, \model first encodes the heterogeneous graph into dense representations. Based on this, the variational inference technique is adopted to learn its stochastic variables. 

\vpara{Heterogeneous attribute representation.} Before feature representation, we randomly mask a portion of the attributes of the target node type, which aim to enhance the model's ability to generalize and handle noisy or incomplete data. We use $\gamma$ to denote the mask rate, $V_t$ denotes the node set of the target type, and the nodes being masked can be denoted as $\tilde{V}$. The attribute of each node in $\tilde{V}$ is assigned with a learnable token, [M]. For each node $v$ in $V_t$, its node attribute can be denoted as:
\[
\widetilde{\boldsymbol{x}}_v= \begin{cases}x_{[M]} & \text { if } v \in \tilde{V} \\ x_v & \text { if } v \notin \widetilde{V}\end{cases}
\]
We use $\tilde{\gH}$ to denote the graph attribute after masking, which will be fed into the encoder for encoding. The mask-then-predict strategy forces the model to rely solely on visible features for training, further enhancing the robustness of the \model.

Based on the masked feature $\tilde{\gH}$, we then adopt HAN~\cite{HAN} as the feature encoder to capture the heterogeneous attributes of the graph. HAN encodes the HINs through node-level and semantic-level attention based on predefined meta-paths. We design various meta-paths tailored to different HINs. For instance, the DBLP graph contains nodes of four different types: author (A), paper (P), conference (C), and term (T) nodes. We aim to forecast the category of author nodes; therefore, we design three meta-paths, namely "APA," "APCPA," and "APTPA," to capture the node-level characteristics of DBLP. The full meta-paths we designed for each graph are listed in Table~\ref{method_metapath}. Based on the pre-designed meta-paths, we adopt HAN to encode each heterogeneous graph as described in Section~\ref{method:han_encode}, resulting in the dense representation $\gH$.

\vpara{Variational inference (VI).}~\label{vi:principle} Based on $\mathcal{H}$, we map it into a latent distribution $p_\theta$, parameterized by $\theta$, to learn its stochastic variables, $Z$. Considering a single point, the probabilistic likelihood between the input embedding $h$ and the latent variable $z$ can be denoted as $p_\theta(h|z)$, and the posterior probability can be denoted as $p_\theta(z|h)$.

As a generative SSL method, \model generates new samples by first sampling $z^{(i)}$ from a prior distribution $p_{\theta^{*}}(z)$, where $\theta^{*}$ represents the optimal parameter for $p$. The distribution $p_{\theta^{*}}(z)$ is assumed to be standard normal. Then, we can generate new samples $\bar{x}$ from the conditional distribution as $p_{\theta^{*}}(\bar{x}|z=z^{(i)})$.

However, as pointed out in VGAE~\cite{VGAE}, it is difficult to compute all possible values of $\mathbf{Z}$. Therefore, we introduce a new approximation function $q_\omega$, parameterized by $\omega$, to generate the approximation of $h$. Following the method in VGAE~\cite{VAE}, we adopt a variational distribution of the following form:

\begin{equation}
q(\mathbf{Z} \mid \mathcal{H}, \mathcal{A})=\prod_{i=1}^N q\left(\mathbf{z}_i \mid \mathcal{H}, \mathcal{A} \right)=\prod_{i=1}^N \mathcal{N}\left(\mathbf{z}_i \mid \boldsymbol{\mu}_i, \boldsymbol{\sigma}_i\right),
\end{equation}
where $\mu_i$ and $\sigma_i$ are the mean and log-variance of $q_{\omega}(z_i)$, and $\mathcal{A}$ represents the adjacency matrix of the target node. In practical implementation, we utilize another two HAN layers to learn $\mu$ and $\sigma$ as follows:

\begin{equation}
\mu = \operatorname{Norm}(\operatorname{HAN}(\mathcal{H}, \mathcal{A})), \quad \sigma = \operatorname{Norm}(\operatorname{HAN}(\mathcal{H}, \mathcal{A})), 
\end{equation}
where $\operatorname{Norm}$ is the normalization operation at the last dimension to avoid extreme values. Based on the learned stochastic latent variable, we adopt the reparameterization technique~\cite{VAE} to generate new samples from the posterior $\mathbf{z}_i \sim q\left(\mathbf{z}_i \mid \mathcal{H}, \mathcal{A}\right)$ as follows:
\begin{equation}
\mathbf{z}_i=\boldsymbol{\mu}_i+\boldsymbol{\sigma}_i \odot \epsilon ; \quad \epsilon \sim \mathcal{N}(0,1),
\label{repam}
\end{equation}
where $\odot$ denotes element-wise multiplication, and $\epsilon$ is a sample from the standard normal distribution. To generate high-quality latent variables and ensure stable training, we constrain the closeness between the approximate posterior $q_{\omega}(z|h)$ and the real posterior $p_{\theta}(z|h)$ with the Kullback-Leibler (KL) divergence. The Evidence Lower Bound (ELBO) is defined as follows:
\begin{equation}
\mathcal{L}_{elbo}=\mathbb{E}_{z \sim q_{\omega}(\mathbf{Z} \mid \mathcal{H}, \mathcal{A})} \log p(\mathcal{H} \mid \mathbf{Z})-\mathrm{D}_{\mathcal{KL}}(q_{\omega}(\mathbf{Z} \mid \mathcal{H}, \mathcal{A}) \| p_{\theta}(\mathbf{Z})).
\label{elbo}
\end{equation}

\subsection{Generative Contrastive Feature Learning}~\label{method:contrastive_learning}
As discussed in Section~\ref{sec:intro}, \model focuses on improving the quality of latent representation through contrastive learning. The critical issue in Graph Contrastive Learning (GCL) lies in the contrastive samples, particularly the selection of hard negative samples. To obtain high-quality contrastive samples, we construct a triple set consisting of the anchor node, positive samples, and negative samples, denoted as $(U, U^{+}, U^{-})$. We will elaborate on the generation process of positive and negative samples respectively.

\vpara{Positive samples generation.} In CV, a common and effective approach is to apply random transformations to the same image to generate positive samples. However, in our work, we adopt a different strategy to generate positive samples, akin to the approach used in SimCSE~\cite{simcse} for generating sentence embeddings. Specifically, we utilize the same heterogeneous encoder to encode the sample input graph twice, resulting in two representations, denoted as $\mathcal{H}_1$ and $\mathcal{H}_2$. Since the attention layer with HAN contains dropout masks on the input features, $\mathcal{H}_1$ and $\mathcal{H}_2$ can be considered a form of data augmentation, differing only in their dropout masks. Thus, we treat $\mathcal{H}_1$ as the anchor sample and $\mathcal{H}_2$ as the positive sample.

\vpara{Progressive negative samples generation.} Unlike previous contrastive methods that synthesize hard negative samples from existing samples~\cite{hardnegative,progcl}, we propose for the first time to leverage the generative ability of \model to generate negative samples. Our aim is to enhance the difficulty of negative samples, thereby compelling the model to better discriminate between similar samples. The key characteristic of negative samples is that they should be semantically similar to the anchor nodes but belong to distinct categories.
 
To achieve this, we propose a novel mechanism to generate negative samples that satisfy the hardness requirement. This mechanism involves negative samples $U^{-}$ from two sources. On one hand, we apply data augmentation on the encoded representation $\mathcal{H}_1$, forming the negative set $U_{D}^{-}$. In $U_{D}^{-}$, each data point is directly generated by applying dropout operation on $\mathcal{H}_1$, denoted as: $u_{d}^{-} = \text{Dropout}(\mathcal{H}_1)$.


On the other hand, the negative samples are generated based on VI, forming $U_{V}^{-}$. For $U_{V}^{-}$, we employ VI with shifted stochastic variables to generate negative samples, as illustrated in Figure~\ref{fig:vi}, the shifted stochastic variables is reassembled with shifted mean value and identical variance. Our approach to generate hard negative samples is simple yet ingenious. Generally, VI first samples a random data point from a Gaussian prior, then utilizes the stochastic variables of mean and variance to generate latent representations using the reparameterization technique~\cite{VGAE}. Given the well-learned stochastic variables, we can generate samples that realistically represent latent data similar to the input data. This phenomenon inspires us to generate latent data as negative samples. Specifically, we introduce a slight shift on the learned stochastic mean variable, to ensure that the generated samples resemble the anchor sample but possess distinct implicit characteristics. We keep the learned variance constant but shift the mean variable with a small coefficient as follows:
\begin{equation}
\boldsymbol{\mu^{*}} = \kappa  * \boldsymbol{\mu},
\end{equation}
where $\kappa$ denotes the coefficient. Following this, we can utilize $\boldsymbol{\mu^{*}}$ and $\boldsymbol{\sigma}$ to generate negative samples as follows:
\begin{equation}
u_{v}^{-} =\boldsymbol{\mu^{*}}+\boldsymbol{\sigma} \odot \epsilon,
\label{nega:2}
\end{equation}
where $\epsilon$ is a random value sampled from the standard normal distribution. As the training progresses, the quality of the generated samples gradually improves and becomes more similar to the anchor samples, while the shifted mean variable $\boldsymbol{\mu^{*}}$ ensures that the generated samples fundamentally belong to a different category than the anchor sample.

We aim to provide continuous hard negative samples during the training process; therefore, the negative samples from $U_{D}^{-}$ and $U_{V}^{-}$ play different roles at different training stages. In the early stage, when the overall training is insufficient, the quality of samples generated by VI is relatively poor, which may not provide enough contrastive negative signal. However, dropout operations with a low dropout ratio can offer necessary negative samples to challenge the model training difficulty at the initial training stage~\cite{dropmessage}. Therefore, the samples of $U_{D}^{-}$ dominate the portion at the early training stage. As training progresses, the quality of samples generated by VI gradually improves, making them more suitable as hard negative samples. We gradually reduce the proportion of $U_{D}^{-}$ and increase the percentage of $ U_{V}^{-}$ over time. This adaptive negative sample generation mechanism generates rather than synthesizes negative samples. Moreover, the combination of negative samples from $U_{D}^{-}$ and $U_{V}^{-}$ ensures the hardness of negative samples, which contributes to contrastive SSL as well as the enhancement of latent representation.

We denote this mechanism as \textbf{\textit{progressive negative samples generation (PNSM)}}, formulated with a balance coefficient $\lambda$ as:
\begin{equation}
U^{-} = \lambda *  U_{D}^{-}  + (1 - \lambda ) * U_{V}^{-}.
\end{equation}
Assuming that the total number of training epochs is $T$, $\lambda$ is calculated as $\lambda = 1 - \frac{t}{T}$ at training step $t$. As the training progresses, $\lambda$ gradually decreases from 1 to 0. This mechanism effectively balances the contribution of both types of negative samples during different stages of training.

\vpara{Contrastive SSL objective.} We adopt InfoNCE~\cite{infonce} as the criterion to measure the divergence between the triple set $(U, U^{+}, U^{-})$. Generally, $U^{+}$ contains a single positive sample and $U^{-}$ contains tens of negative samples. For an anchor node $u$ and positive node $v$, the loss function can be defined as:

\begin{equation}
    \mathcal{L}_{pnsm}=-\log \frac{\exp \left(\operatorname{sim}\left(h_{u}, h_{v^{+}}\right) / \tau\right)}{\sum_{k=1, k \neq i}^{U^{-}} \exp \left(\operatorname{sim}\left(h_{u}, h_{k}\right) / \tau\right)},
\label{info}
\end{equation}
where $\tau$ denotes the temperature parameter.
 
\subsection{Decoding}~\label{method:decoding}
The feature reconstruction criterion varies for different graph tasks~\cite{VGAE,GraphMAE}. The most common reconstruction function is mean square error (MSE)~\cite{mae}. However, MSE assumes data independence and ignores the spatial relationships of samples~\cite{mseshortcoming}, whereas graph data is inherently connected. Besides, as pointed out in GraphMAE~\cite{GraphMAE}, MSE suffers issues of sensitivity and low selectivity, which may resulting in overfitting caused by extreme data points. To address this, GraphMAE~\cite{GraphMAE} proposes using scaled cosine error (SCE) as the reconstruction criterion. SCE aims to down-weight easy samples to force the model to focus on samples that are difficult to distinguish. It scales the cosine error with a power factor $\delta$ larger than 1. In addition, we notice that Focal Loss~\cite{focalloss} is commonly leveraged to address classification problems with hard samples, formulated as:
\begin{equation}
\mathrm{FL}(p)=-\left(1-p\right)^\delta \log(p),
\end{equation}
where $p$ is the estimated probability. When a sample is misclassified and $p$ is small, $1-p$ is near 1 and the loss won't be affected. When $p$ approaches 1, $1-p$ is small and the loss goes to 0, in this way, well-classified samples are down-weighted.
 
Motivated by SCE and Focal Loss, we propose an enhanced scaled cosine error (ESCE) to reconstruct the input graph feature. In comparison to SCE, ESCE enhances the original formula by adding a $\log$ term. This modification brings advantage in tackling difficult reconstruction samples. ESCE is formally consistent with focal loss, which allows the model to pay more attention to nodes that are difficult to reconstruct. To ensure numerical stability during training, since the cosine similarity values range between $[-1, 1]$, we set negative similarity values to an extremely small number, e.g., 1e-6, to avoid possible overflow in logarithmic calculations. ESCE takes the form:
\begin{small}
\begin{equation}
\mathcal{L}_{\mathrm{esce}}=\frac{1}{|\widetilde{\mathcal{V}}|} \sum_{v_i \in \widetilde{\mathcal{V}}}\left(1-\frac{x_i^T z_i}{\left\|x_i\right\| \cdot\left\|z_i\right\|}\right)^\delta \log\left(1-\frac{x_i^T z_i}{\left\|x_i\right\| \cdot\left\|z_i\right\|}\right) , \delta \geq 1.
\label{esce}
\end{equation}
\end{small}

\subsection{Training Objectives}~\label{method:training_objective}
The overall training objective is defined as a weighted combination of the ELBO $\mathcal{L}_{elbo}$, the contrastive loss $\mathcal{L}_{pnsm}$, and the reconstruction loss $\mathcal{L}_{esce}$:
\begin{equation}
\mathcal{L} = \alpha\mathcal{L}_{elbo} + \beta\mathcal{L}_{pnsm} + \gamma\mathcal{L}_{esce},
\end{equation}
where $\alpha$, $\beta$, and $\gamma$ are intermediary weights used to balance $\mathcal{L}_{elbo}$, $\mathcal{L}_{pnsm}$, and $\mathcal{L}_{esce}$, respectively. The overall training process is summarized in Appendix~\ref{algor:training}. Additionally, we compare \model with three of the most related works, and the detailed comparative analysis is presented in Appendix~\ref{method_comparison}.

\section{EXPERIMENTS}
We conduct extensive experiments with the aim of addressing the following three research questions (\textbf{RQ}) in this section: \textbf{RQ1}: How does \model perform compared to other state-of-the-art HGL methods? \textbf{RQ2}: How do different training strategies affect the overall performance? \textbf{RQ3}: How do different parameters of \model affect the overall performance?
 
We begin by introducing the datasets, baselines, and experimental settings, which can be found in Appendix~\ref{apx:experiment_settings}. Then we compare \model with comprehensive SOTA baselines of 4 types on the tasks of node classification and node clustering. Following this, we conduct ablation studies to investigate the effect of diverse training strategies, followed by a parameter sensitivity analysis. Finally, we perform embedding visualization to provide an intuitive assessment of the quality of representations learned by various baselines.

\subsection{Main Results}
We conduct experiments on node-level tasks to answer \textbf{RQ1}.

\vpara{Results of node classification.} We utilize the learned node representations to train linear classifiers for node-level tasks. For a comprehensive evaluation, we vary the number of labeled nodes per class in the training set (20, 40, and 60), while keeping the validation and test sets consistent with 1000 nodes each across all datasets. Test performance is reported based on the best result obtained on the validation set, using Micro-F1 and Macro-F1 as evaluation metrics. 

Table~\ref{main:classification} presents the results, with the best performance highlighted in bold. Compared to other SOTA baselines of four categories, \model achieve the best in 18 out of 24 experiments, securing the second-best results in the rest 6. We can draw three conclusions from the results. First, despite that the datasets exhibit diverse characteristics, \model demonstrates its capability to achieve high performance across diverse graph datasets, highlighting its superiority and robustness. Second, it is observed that \model exhibits lower variance compared to other methods, indicating its stability. Third, generative models generally outperform methods of the other three types, albeit generative SSL still lacks extensive exploration. Hence, these findings collectively validate the effectiveness of \model and underscore the potential of generative models.

\begin{table*}[]
\centering
\caption{Results of node classification.}
\resizebox{\linewidth}{!}{
\begin{tabular}{@{}cccccccccccccc>{\columncolor{mygrey}}c@{}}
\toprule
\multicolumn{3}{c|}{}                                                           & \multicolumn{2}{c|}{Network Embedding}  & \multicolumn{3}{c|}{HGNNs}                       & \multicolumn{3}{c|}{Contrastive SSL}              & \multicolumn{3}{c|}{Genreative SSL}                & Ours               \\ \midrule
Datasets                  & Metric                 & \multicolumn{1}{c|}{Split} & GraphSAGE & \multicolumn{1}{c|}{Mp2vec} & HERec     & HetGNN    & \multicolumn{1}{c|}{HAN} & DGI       & DMGI      & \multicolumn{1}{c|}{HeCo} & GAE       & GraphMAE  & \multicolumn{1}{c|}{HGMAE} & \model             \\ \midrule
\multirow{6}{*}{DBLP}     & \multirow{3}{*}{Mi-F1} & 20                         & 71.44±8.7 & 89.67±0.1                   & 90.24±0.4 & 90.11±1.0 & 90.16±0.9                & 88.72±2.6 & 90.78±0.3 & \underline{91.97±0.2}                 & 91.55±0.1 & 89.31±0.7 & 91.74±0.6                  & \textbf{93.20±0.1} \\
                          &                        & 40                         & 73.61±8.6 & 89.14±0.2                   & 90.15±0.4 & 89.03±0.7 & 89.47±0.9                & 89.22±0.5 & 89.92±0.4 & 90.76±0.3                 & 90.00±0.3 & 87.80±0.5 & \textbf{91.97±0.4}         & \underline{91.68±0.0}          \\
                          &                        & 60                         & 74.05±8.3 & 91.17±0.1                   & 91.01±0.3 & 90.43±0.6 & 90.34±0.8                & 90.35±0.8 & 90.66±0.5 & 91.59±0.2                 & 90.95±0.2 & 89.82±0.4 & \underline{92.76±0.4}                  & \textbf{94.08±0.0} \\ \cmidrule(l){2-15} 
                          & \multirow{3}{*}{Ma-F1} & 20                         & 71.97±8.4 & 88.98±0.2                   & 89.57±0.4 & 89.51±1.1 & 89.31±0.9                & 87.93±2.4 & 89.94±0.4 & \underline{91.28±0.2}                 & 90.90±0.1 & 87.94±0.7 & 91.14±0.6                  & \textbf{92.82±0.1} \\
                          &                        & 40                         & 73.69±8.4 & 88.68±0.2                   & 89.73±0.4 & 88.61±0.8 & 88.87±1.0                & 88.62±0.6 & 89.25±0.4 & 90.34±0.3                 & 89.60±0.3 & 86.85±0.7 & \underline{91.62±0.4}                  & \textbf{91.68±0.1} \\
                          &                        & 60                         & 73.86±8.1 & 90.25±0.1                   & 90.18±0.3 & 89.56±0.5 & 89.20±0.8                & 89.19±0.9 & 89.46±0.6 & 90.64±0.3                 & 90.08±0.2 & 88.07±0.6 & \underline{92.03±0.5}                  & \textbf{93.39±0.1} \\ \midrule
\multirow{6}{*}{Freebase} & \multirow{3}{*}{Mi-F1} & 20                         & 54.83±3.0 & 56.23±0.8                   & 57.92±0.5 & 56.85±0.9 & 57.24±3.2                & 58.16±0.9 & 58.26±0.9 & 61.72±0.6                 & 55.20±0.7 & \underline{64.88±1.8} & 64.18±1.0                  & \textbf{67.32±2.2} \\
                          &                        & 40                         & 57.08±3.2 & 61.01±1.3                   & 62.71±0.7 & 53.96±1.1 & 63.74±2.7                & 57.82±0.8 & 54.28±1.6 & 64.03±0.7                 & 56.05±2.0 & 62.34±1.0 & \textbf{67.82±0.7}         & \underline{66.72±1.2}          \\
                          &                        & 60                         & 55.92±3.2 & 58.74±0.8                   & 58.57±0.5 & 56.84±0.7 & 61.06±2.0                & 57.96±0.7 & 56.69±1.2 & 63.61±1.6                 & 53.85±0.4 & 59.48±6.2 & \underline{66.51±0.8}                 & \textbf{67.32±1.2} \\ \cmidrule(l){2-15} 
                          & \multirow{3}{*}{Ma-F1} & 20                         & 45.14±4.5 & 53.96±0.7                   & 55.78±0.5 & 52.72±1.0 & 53.16±2.8                & 54.90±0.7 & 55.79±0.9 & 59.23±0.7                 & 53.81±0.6 & 59.04±1.0 & \underline{61.34±0.8}                 & \textbf{62.72±1.6} \\
                          &                        & 40                         & 44.88±4.1 & 57.80±1.1                   & 59.28±0.6 & 48.57±0.5 & 59.63±2.3                & 53.40±1.4 & 49.88±1.9 & 61.19±0.6                 & 52.44±2.3 & 56.40±1.1 & \textbf{64.84±0.7}         & \underline{61.97±0.8}        \\
                          &                        & 60                         & 45.16±3.1 & 55.94±0.7                   & 56.50±0.4 & 52.37±0.8 & 56.77±1.7                & 53.81±1.1 & 52.10±0.7 & 60.13±1.3                 & 50.65±0.4 & 51.73±2.3 & \textbf{63.23±0.7}         & \underline{62.49±0.8}          \\ \midrule
\multirow{6}{*}{ACM}      & \multirow{3}{*}{Mi-F1} & 20                         & 49.72±5.5 & 53.13±0.9                   & 57.47±1.5 & 71.89±1.1 & 85.11±2.2                & 79.63±3.5 & 87.60±0.8 & 88.13±0.8                 & 68.02±1.9 & 82.48±1.9 & \underline{87.95±1.0}                 & \textbf{88.62±0.3} \\
                          &                        & 40                         & 60.98±3.5 & 64.43±0.6                   & 62.62±0.9 & 74.46±0.8 & 87.21±1.2                & 80.41±3.0 & 86.02±0.9 & 87.45±0.5                 & 66.38±1.9 & 82.93±1.1 & \underline{89.33±0.4}                  & \textbf{90.40±0.3} \\
                          &                        & 60                         & 60.72±4.3 & 62.72±0.3                   & 65.15±0.9 & 76.08±0.7 & 88.10±1.2                & 80.15±3.2 & 87.82±0.5 & 88.71±0.5                 & 65.71±2.2 & 80.77±1.1 & \underline{89.10±0.4}                  & \textbf{92.10±0.0} \\ \cmidrule(l){2-15} 
                          & \multirow{3}{*}{Ma-F1} & 20                         & 47.13±4.7 & 51.91±0.9                   & 55.13±1.5 & 72.11±0.9 & 85.66±2.1                & 79.27±3.8 & 87.86±0.2 & 88.56±0.8                 & 62.72±3.1 & 82.26±1.5 & \underline{88.83±0.4}                  & \textbf{89.25±0.6} \\
                          &                        & 40                         & 55.96±6.8 & 62.41±0.6                   & 61.21±0.8 & 72.02±0.4 & 87.47±1.1                & 80.23±3.3 & 86.23±0.8 & 87.61±0.5                 & 61.61±3.2 & 82.00±1.1 & \underline{89.39±0.4}                  & \textbf{90.26±0.4} \\
                          &                        & 60                         & 56.59±5.7 & 61.13±0.4                   & 64.35±0.8 & 74.33±0.6 & 88.41±1.1                & 80.03±3.3 & 87.97±0.4 & 89.04±0.5                 & 61.67±2.9 & 80.29±1.0 & \underline{89.32±0.3}                  & \textbf{92.22±0.0} \\ \midrule
\multirow{6}{*}{Aminer}   & \multirow{3}{*}{Mi-F1} & 20                         & 49.68±3.1 & 60.82±0.4                   & 63.64±1.1 & 61.49±2.5 & 68.86±4.6                & 62.39±3.9 & 63.93±3.3 & 78.81±1.3                 & 65.78±2.9 & 68.21±0.3 & \underline{80.76±0.9}                  & \textbf{80.98±0.1} \\
                          &                        & 40                         & 52.10±2.2 & 69.66±0.6                   & 71.57±0.7 & 68.47±2.2 & 76.89±1.6                & 63.87±2.9 & 63.60±2.5 & 80.53±0.7                 & 71.34±1.8 & 74.23±0.2 & \textbf{82.84±1.0}         & \underline{81.32±0.1}          \\
                          &                        & 60                         & 51.36±2.2 & 63.92±0.5                   & 69.76±0.8 & 65.61±2.2 & 74.73±1.4                & 63.10±3.0 & 62.51±2.6 & 82.46±1.4                 & 67.70±1.9 & 72.28±0.2 & \underline{82.11±0.7}                  & \textbf{82.66±0.2} \\ \cmidrule(l){2-15} 
                          & \multirow{3}{*}{Ma-F1} & 20                         & 42.46±2.5 & 54.78±0.5                   & 58.32±1.1 & 50.06±0.9 & 56.07±3.2                & 51.61±3.2 & 59.50±2.1 & 71.38±1.1                 & 60.22±2.0 & 62.64±0.2 & \underline{72.79±1.1}                & \textbf{72.86±0.1} \\
                          &                        & 40                         & 45.77±1.5 & 64.77±0.5                   & 64.50±0.7 & 58.97±0.9 & 63.85±1.5                & 54.72±2.6 & 61.92±2.1 & 73.75±0.5                 & 65.66±1.5 & 68.17±0.2 & \textbf{75.52±1.1}                  & \underline{74.87±0.1} \\
                          &                        & 60                         & 44.91±2.0 & 60.65±0.3                   & 65.53±0.7 & 57.34±1.4 & 62.02±1.2                & 55.45±2.4 & 61.15±2.5 & 75.80±1.8                 & 63.74±1.6 & 68.21±0.2 & \underline{75.50±1.0}                  & \textbf{76.43±0.1} \\ \bottomrule
\end{tabular}}
\label{main:classification}
\end{table*}

\vpara{Results of Node Clustering.} Furthermore, we conduct experiments on the node clustering task and present the results in Table~\ref{res:clustering}. \model achieves comparable performance with the other 10 baselines, securing the best results in 6 out of 8 experiments and the second-best results in the remaining three experiments. The performance comparison between HGMAE and \model indicates a relatively close performance, reaffirming the potential of generative models in heterogeneous graph learning.

\begin{table}[htb]
\centering
\caption{Results of node clustering.}
\resizebox{\linewidth}{!}{
\begin{tabular}{ccccccccc}
\hline
          & \multicolumn{2}{c}{DBLP}        & \multicolumn{2}{c}{Freebase}    & \multicolumn{2}{c}{ACM}         & \multicolumn{2}{c}{AMiner}      \\ \hline
Metrics   & NMI            & ARI            & NMI            & ARI            & NMI            & ARI            & NMI            & ARI            \\ \hline
GraphSAGE & 51.5           & 36.4           & 9.05           & 10.49          & 29.2           & 27.72          & 15.74          & 10.1           \\
GAE       & 72.59          & 77.31          & 19.03          & 14.1           & 27.42          & 24.49          & 28.58          & 20.9           \\
Mp2vec    & 73.55          & 77.7           & 16.47          & 17.32          & 48.43          & 34.65          & 30.8           & 25.26          \\
HERec     & 70.21          & 73.99          & 19.76          & 19.36          & 47.54          & 35.67          & 27.82          & 20.16          \\
HetGNN    & 69.79          & 75.34          & 12.25          & 15.01          & 41.53          & 34.81          & 21.46          & 26.6           \\
DGI       & 59.23          & 61.85          & 18.34          & 11.29          & 51.73          & 41.16          & 22.06          & 15.93          \\
DMGI      & 70.06          & 75.46          & 16.98          & 16.91          & 51.66          & 46.64          & 19.24          & 20.09          \\
HeCo      & 74.51          & 80.17          & 20.38          & 20.98          & 56.87          & 56.94          & 32.26          & 28.64          \\
GraphMAE  & 65.86          & 69.75          & 19.43          & 20.05          & 47.03          & 46.48          & 17.98          & 21.52          \\
HGMAE     & \textbf{76.92} & \textbf{82.34} & 22.05 & 22.84 & 66.68          & 71.51          & 41.1           & 38.27          \\ \hline
\rowcolor{mygrey} \model    & 75.92          & 80.99          & \textbf{22.52}          & \textbf{23.1}          & \textbf{67.26} & \textbf{72.09} & \textbf{41.38} & \textbf{56.00} \\ \hline
\end{tabular}}
\label{res:clustering}
\end{table}

\subsection{Ablation Study}~\label{sec:abla}
We conduct ablation studies from two aspects to answer \textbf{RQ2}. (1) The effect of the three training strategies on the performance of \model. (2) The effect of different methods for constructing negative samples on model performance.

\vpara{Effect of training strategies.} We sequentially remove one of the three training strategies, allowing only the remaining two strategies to participate in training. The ablation results on DBLP are reported in Table~\ref{exp:abla0_dblp}, the full ablation stduies on four datasets are reported in Appendix~\ref{apx:ablation_studies_training_losses}. Two observations can be made from the results. \textbf{First}, the three losses generally contribute to the final performance of \model across different data partitions. Furthermore, in all ablation experiments except on Freebase, where the Macro-F1 value slightly outperforms the performance of \model after the removal of the reconstruction loss in an experimental setting involving 40 data divisions for each node type, eliminating a specific loss does not result in improved model performance. This confirms the overall effectiveness of feature reconstruction with ESCE. \textbf{Second}, we found that among the three strategies, the contrastive SSL strategy exhibited the greatest impact on the ultimate performance, especially on Freebase, ACM, and AMiner. On AMiner, the model's performance without the contrastive strategy encountered an average decrease of 31.25 in Micro-F1 and an average decrease of 27.48 in Macro-F1 values. This phenomenon validates that contrastive learning on latent representation can greatly improve the overall performance, demonstrating its salient contribution.

\begin{table}[]
\caption{The impact of three training losses on DBLP.}
\begin{tabular}{@{}cccc@{}}
\toprule
Split                & Datasets        & \multicolumn{2}{c}{DBLP}                \\ \midrule
\multicolumn{1}{l}{} & Metrics         & Mi-F1              & Ma-F1              \\ \midrule
\multirow{4}{*}{20}  & w.o. $L_{pnsm}$ & 89.16± 0.9         & 88.46±0.1          \\
                     & w.o. $L_{elbo}$ & 92.30±0.2          & 91.86±0.2          \\
                     & w.o. $L_{esce}$ & 91.88±0.1          & 92.20±0.0          \\
                     & \textbf{\model} & \textbf{92.62±0.1} & \textbf{92.98±0.1} \\ \midrule
\multirow{4}{*}{40}  & w.o. $L_{pnsm}$ & 88.74±1.0          & 88.30±0.9          \\
                     & w.o. $L_{elbo}$ & 88.38±0.0          & 88.98±0.0          \\
                     & w.o. $L_{esce}$ & 89.34±0.0          & 89.78±0.0          \\
                     & \textbf{\model} & \textbf{91.86±0.2} & \textbf{91.65±0.1} \\ \midrule
\multirow{4}{*}{60}  & w.o. $L_{pnsm}$ & 89.66±0.9          & 88.54±0.9          \\
                     & w.o. $L_{elbo}$ & 92.00±0.0          & 91.19±0.0          \\
                     & w.o. $L_{esce}$ & 92.10±0.0          & 91.29±0.0          \\
                     & \textbf{\model} & \textbf{94.04±0.1} & \textbf{93.37±0.1} \\ \bottomrule
\end{tabular}
\label{exp:abla0_dblp}
\end{table}

\vpara{Effect of the negative samples construction.} To explore the impact of PNSM on \model, we develop four variants of \model, each equipped with a different negative sample construction method as: 
\begin{itemize}
    \item $\operatorname{\model}_1$: Samples random noise from a Gaussian prior as negative samples.
    \item $\operatorname{\model}_2$: Utilizes only $U_{D}^{-}$ as negative samples. The samples collected from the encoded representation with dropout.
    \item $\operatorname{\model}_3$: Utilizes only $U_{V}^{-}$ as negative samples, i.e., samples collected using variational inference.
    \item $\operatorname{\model}_4$: Utilizes the same stochastic variance to generate negative samples, i.e., the mean value in Equation~\ref{nega:2} is replaced with $\mu$.
\end{itemize}

\begin{table}[]
\caption{The impact of negative samples generation methods on DBLP.}
\begin{tabular}{@{}cccc@{}}
\toprule
Split               & Datasets                  & \multicolumn{2}{c}{DBLP}                \\ \midrule
                    & Metrics                   & Mi-F1              & Ma-F1              \\ \midrule
\multirow{5}{*}{20} & $\operatorname{\model}_0$ & 88.64±1.1          & 88.68±1.5          \\
                    & $\operatorname{\model}_1$ & 90.94±0.1          & 90.55±0.1          \\
                    & $\operatorname{\model}_2$ & 87.62±2.5          & 84.81±4.4          \\
                    & $\operatorname{\model}_3$ & 92.18±0.2          & 91.79±0.2          \\
                    & \textbf{\model}           & \textbf{92.62±0.2} & \textbf{92.98±0.1} \\ \midrule
\multirow{5}{*}{40} & $\operatorname{\model}_0$ & 88.34±0.6          & 87.45±0.9          \\
                    & $\operatorname{\model}_1$ & 89.02±0.0          & 88.75±0.0          \\
                    & $\operatorname{\model}_2$ & 87.62±0.2          & 86.48±0.3          \\
                    & $\operatorname{\model}_3$ & 89.62±0.0          & 89.41±0.1          \\
                    & \textbf{\model}           & \textbf{91.86±0.2} & \textbf{91.65±0.1} \\ \midrule
\multirow{5}{*}{60} & $\operatorname{\model}_0$ & 89.46±0.4          & 87.30±0.7          \\
                    & $\operatorname{\model}_1$ & 90.70±0.1          & 90.02±0.1          \\
                    & $\operatorname{\model}_2$ & 89.46±0.1          & 87.39±0.1          \\
                    & $\operatorname{\model}_3$ & 91.42±0.0          & 90.66±0.1          \\
                    & \textbf{\model}           & \textbf{94.04±0.1} & \textbf{93.37±0.1} \\ \bottomrule
\end{tabular}
\label{exp:abla_negative_construction_dblp}
\end{table}
 
We present the ablation results of DBLP in Table~\ref{exp:abla_negative_construction_dblp}, the full ablation results on four datasets can be found in Appendix~\ref{apx:ablation_studies_negative_construction}. Three observations can be drawn. \textbf{First}, employing random noise as negative samples generally deteriorates model performance, which suggests that relying solely on high-quality positive samples is insufficient. \textbf{Second}, relying solely on $U_{D}^{-}$ or $U_{V}^{-}$ does not lead to satisfactory performance. Moreover, depending solely on $U_{V}^{-}$ has a more pronounced impact on model performance, potentially due to the inaccurate latent variance learned during the early training phase. Although the model may gradually generate high-quality negative samples in later training stages, it might struggle to optimize in the correct direction. \textbf{Third}, generating negative samples with identical stochastic variables poses issues. These negative samples resemble the anchor sample, making it challenging to distinguish between them and resulting in false negatives.

The above observations demonstrate that PNSM appears to be a more favorable approach for model training. It circumvents the issues encountered in conventional negative sample generation methods, thereby leads to improved performance.

\subsection{Parameter Analysis}
In this subsection, we conduct experiments to address \textbf{RQ3}. The experiments encompass variations in the hidden dimension of the encoder, the shift factor $\kappa$, and the number of negative samples. Due to space limitations, we solely report the results using Micro-F1.

\vpara{Impact of hidden dimension.} We present the impact of the hidden dimension in Appendix~\ref{apx:para_analysis_hidden_dim}, with the hidden dimension varied from \{64, 128, 256, 512, 1024\}. Generally, increasing the hidden dimension enhances performance, likely due to the complexity inherent in non-Euclidean graph data, higher-dimensional variables are necessary to capture intricate relationships within the graph. However, we observe that further increasing the hidden dimension to 512 leads to performance degradation for DBLP and ACM. We believe that a wider hidden layer may deter the model from focusing on meaningful information. Moreover, we note that the optimal hidden dimension can vary across datasets.

\vpara{Impact of mean shift factor $\kappa$.} As revealed in Section~\ref{sec:abla}, the hardness of negative samples is crucial in PNSM, which can be controlled by adjusting the parameter $\kappa$ in \model. We explore values of $\kappa$ within the set \{1, 2, 3, 4, 5, 6\} and present the results in Appendix~\ref{apx:para_analysis_mean_shift}. We can draw three conclusions from it. First, the optimal $\kappa$ value varies across different datasets, indicating that the choice of $\kappa$ should be tailored to the specific dataset to achieve optimal performance. Second, the model's performance fluctuates for different $\kappa$ values, underscoring the impact of this parameter on the quality of negative samples. Third, in general, the performance benefits from relatively small $\kappa$ values, further validating that negative samples should be semantically close to the anchor nodes to effectively improve the model's discriminative ability.

\vpara{Impact of the number of negative samples.} We further conducted experiments to explore the impact of negative sample size on performance. Specifically, we increased the number of negative samples from 5 to 40 and conducted experiments on four datasets to examine the effect. As shown in Appendix~\ref{apx:para_analysis_negative_samples}, the optimal negative sample size varies across datasets. Increasing the number of negative samples generally improves performance; however, the performance improvement reaches a plateau when the number of negative samples is increased to a certain threshold. We attribute this phenomenon to the PNSM strategy, which progressively increases the ratio of hard-negative samples in the generated negative samples. This implies that with a sufficient number of challenging negative samples, excellent performance can be achieved even with a relatively modest number of negative samples.

\subsection{Embedding Visualization}
\begin{figure}[t] 
  \begin{minipage}{0.32\linewidth}
    \vspace{3pt}        
    \centerline{\includegraphics[width=\textwidth]{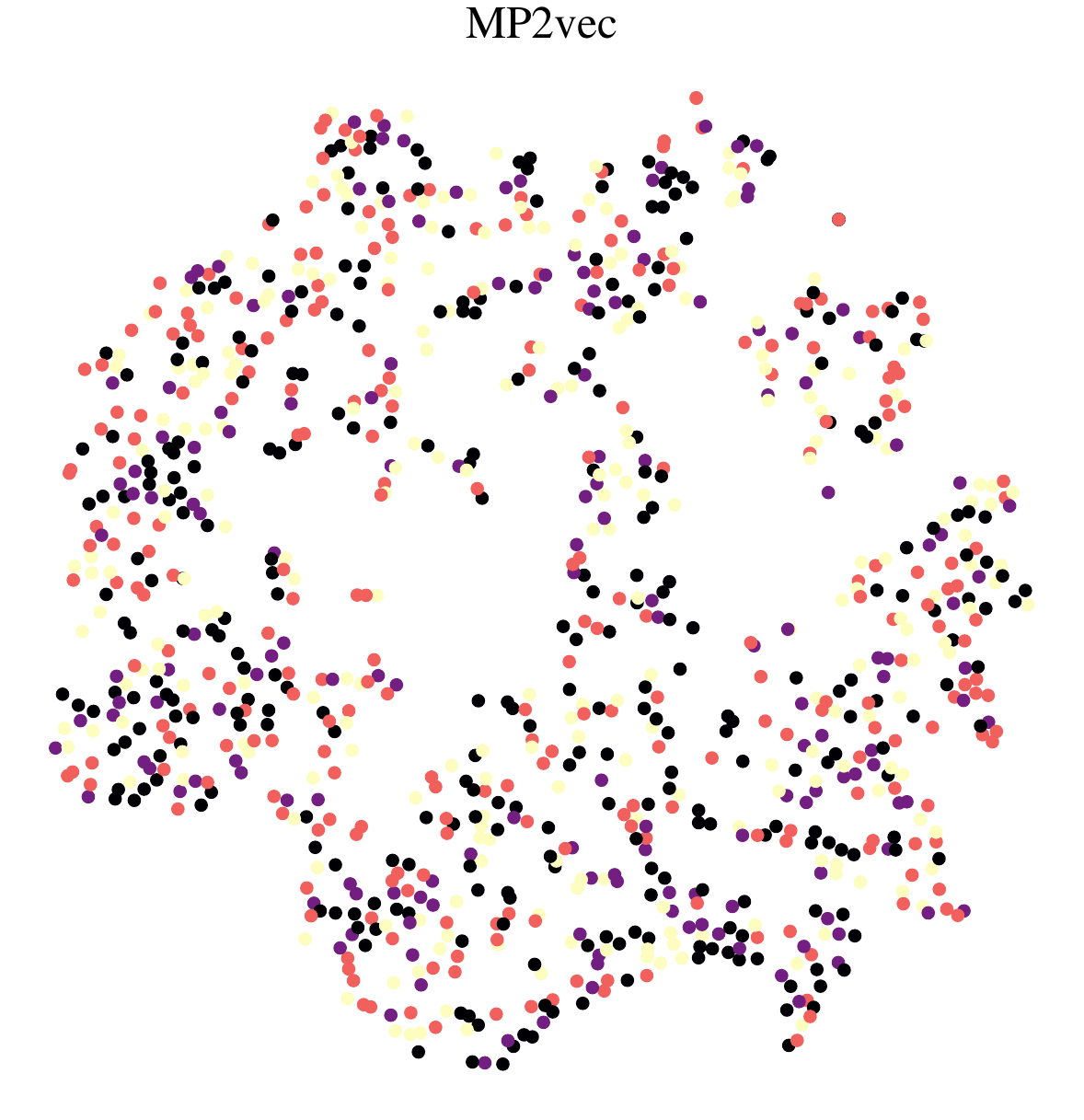}}          
  \end{minipage}
  \begin{minipage}{0.32\linewidth}
    \vspace{3pt}
            \centerline{\includegraphics[width=\textwidth]{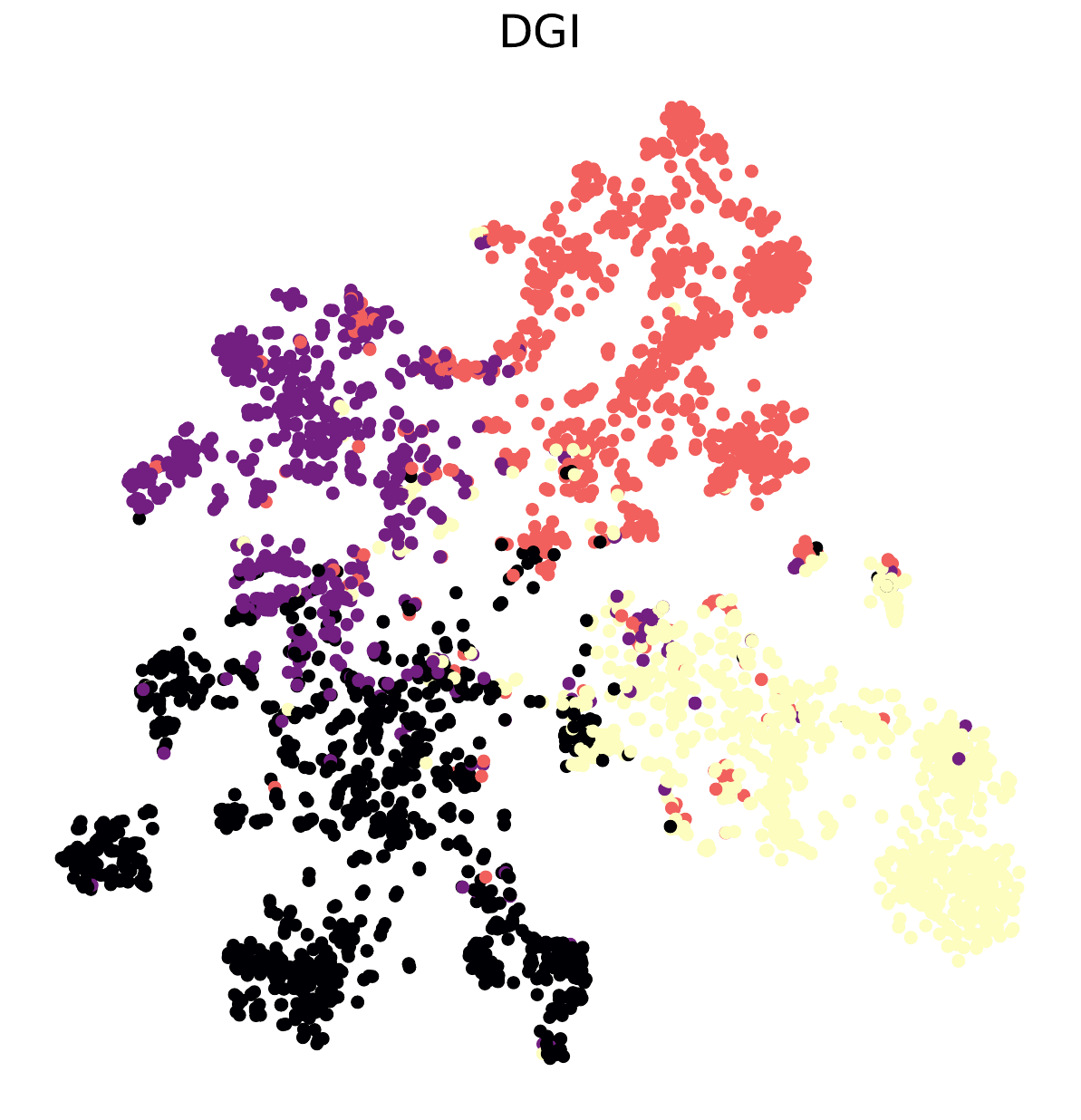}}   
  \end{minipage}
  \begin{minipage}{0.32\linewidth}
    \vspace{3pt}
    \centerline{\includegraphics[width=\textwidth]{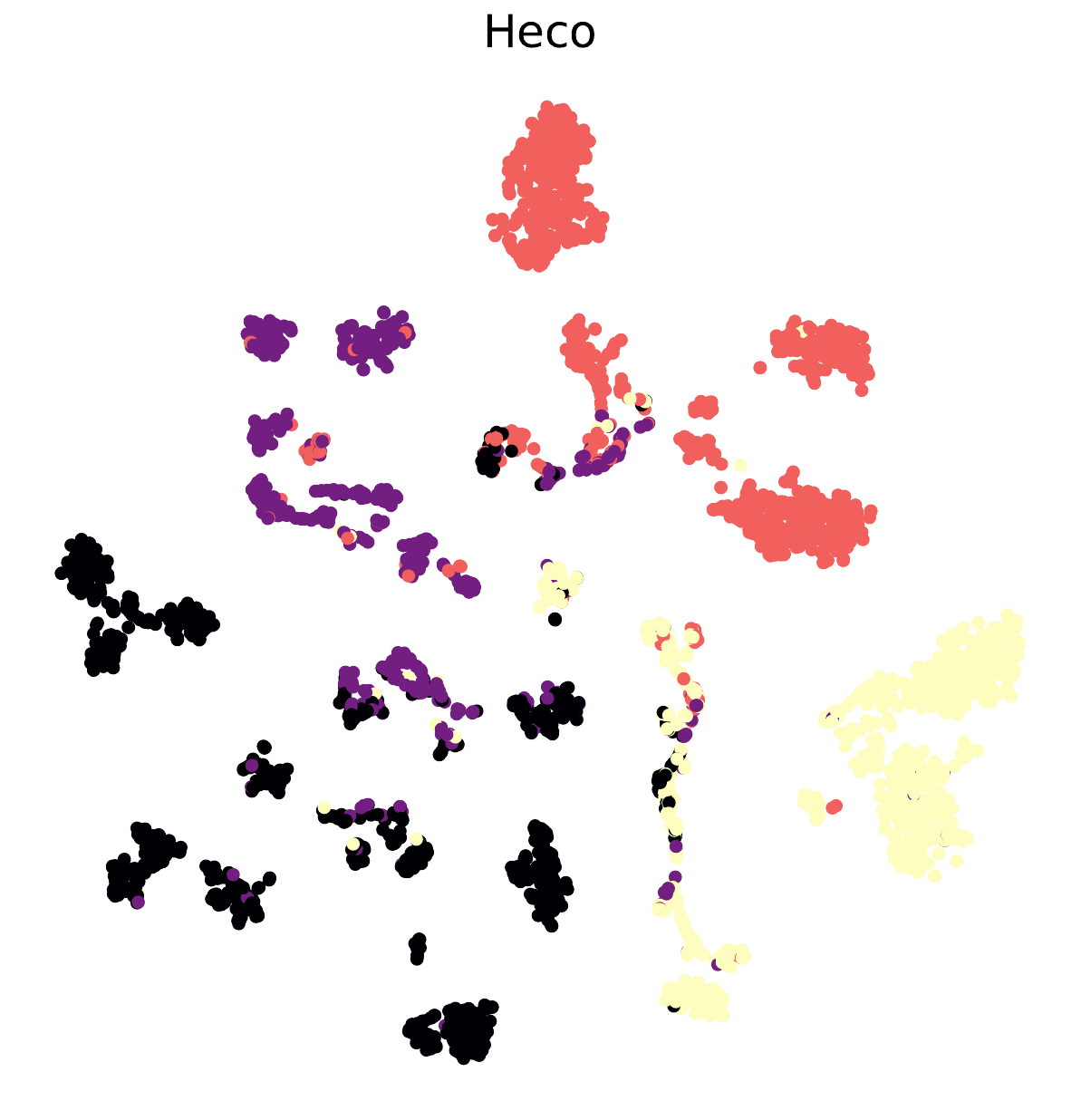}}
  \end{minipage} \\
        \begin{minipage}{0.32\linewidth}
    \vspace{3pt}        
    \centerline{\includegraphics[width=\textwidth]{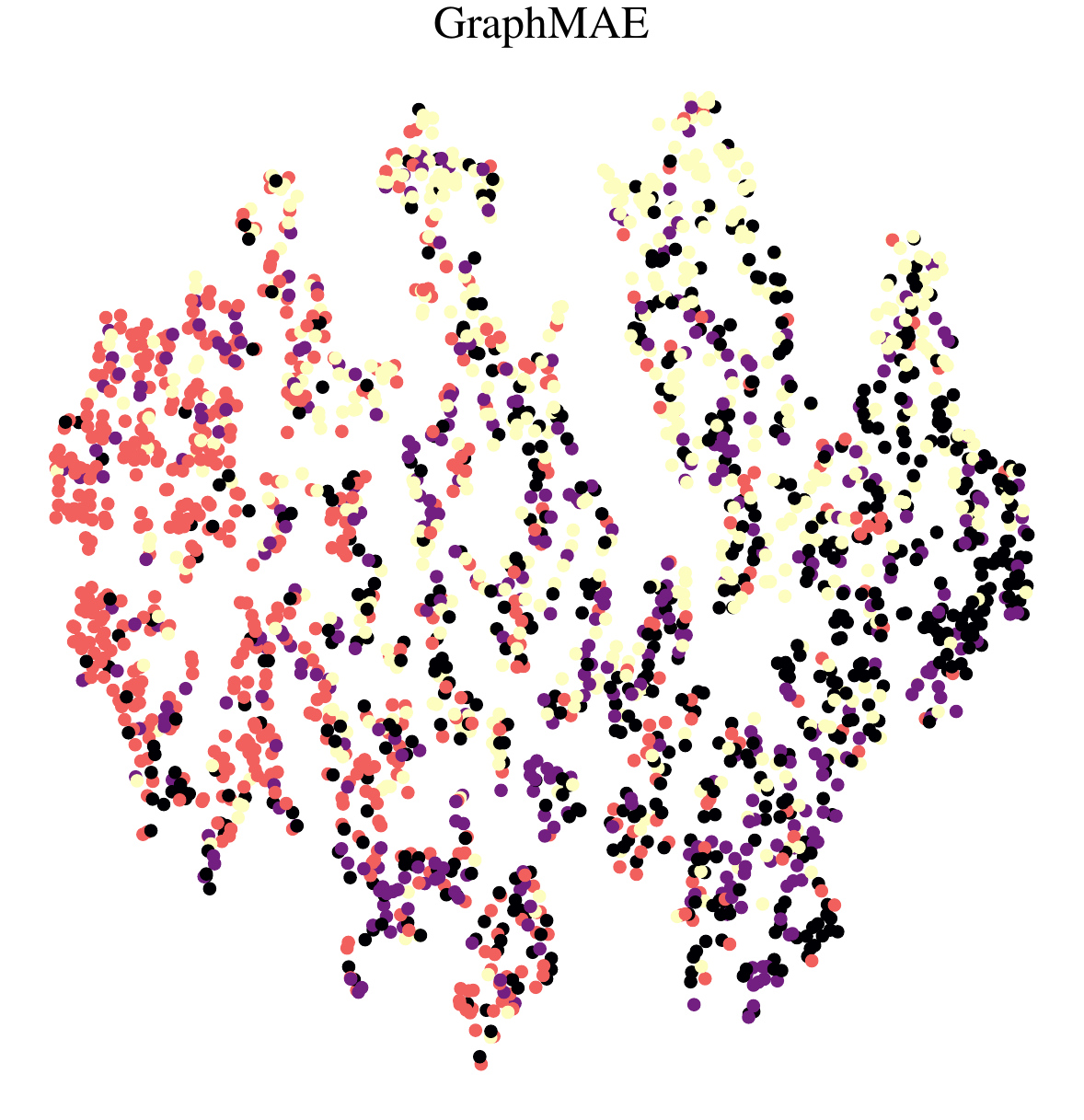}}          
  \end{minipage}
  \begin{minipage}{0.32\linewidth}
    \vspace{3pt}
            \centerline{\includegraphics[width=\textwidth]{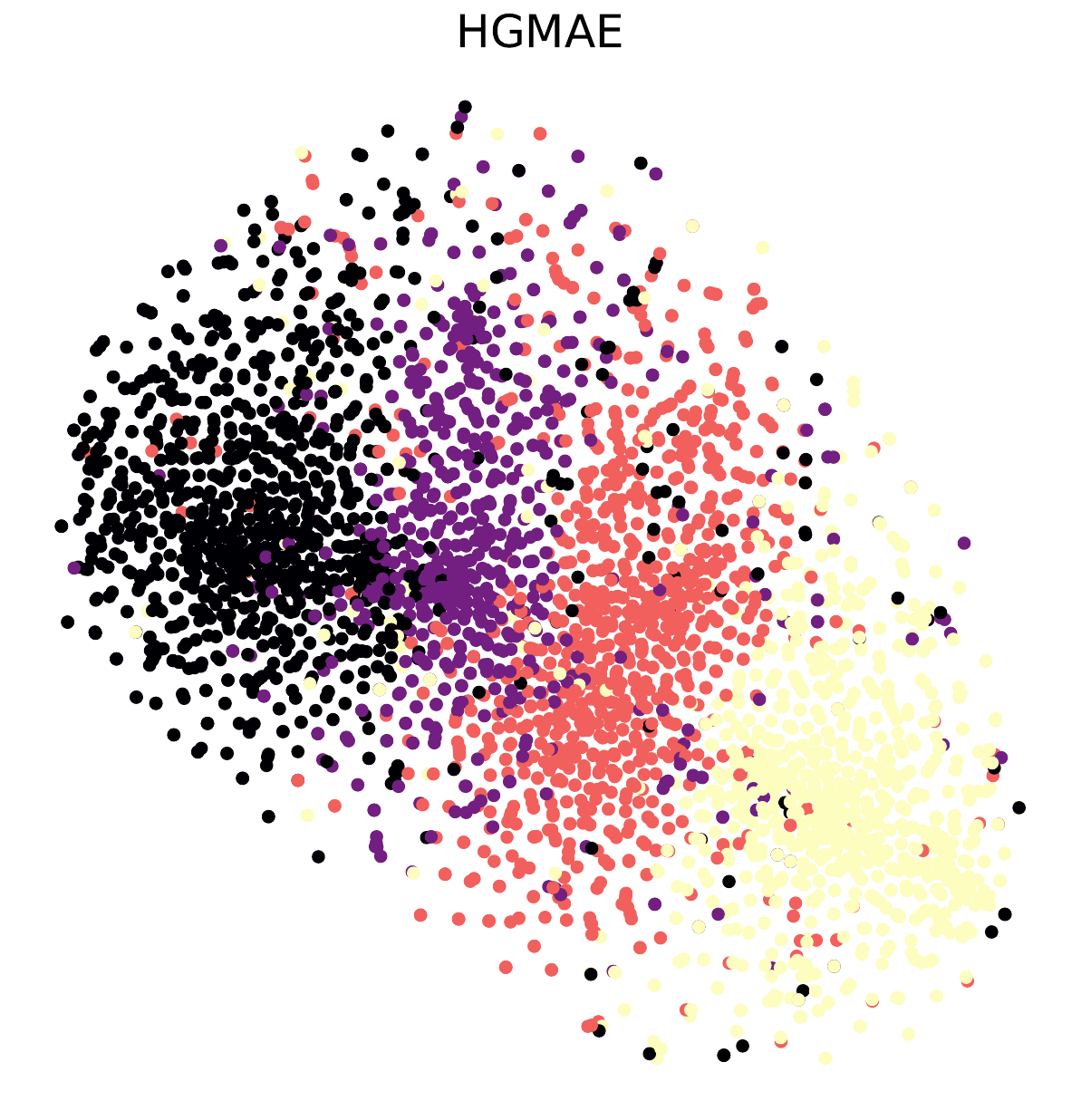}}
  \end{minipage}
  \begin{minipage}{0.32\linewidth}
    \vspace{3pt}
    \centerline{\includegraphics[width=\textwidth]{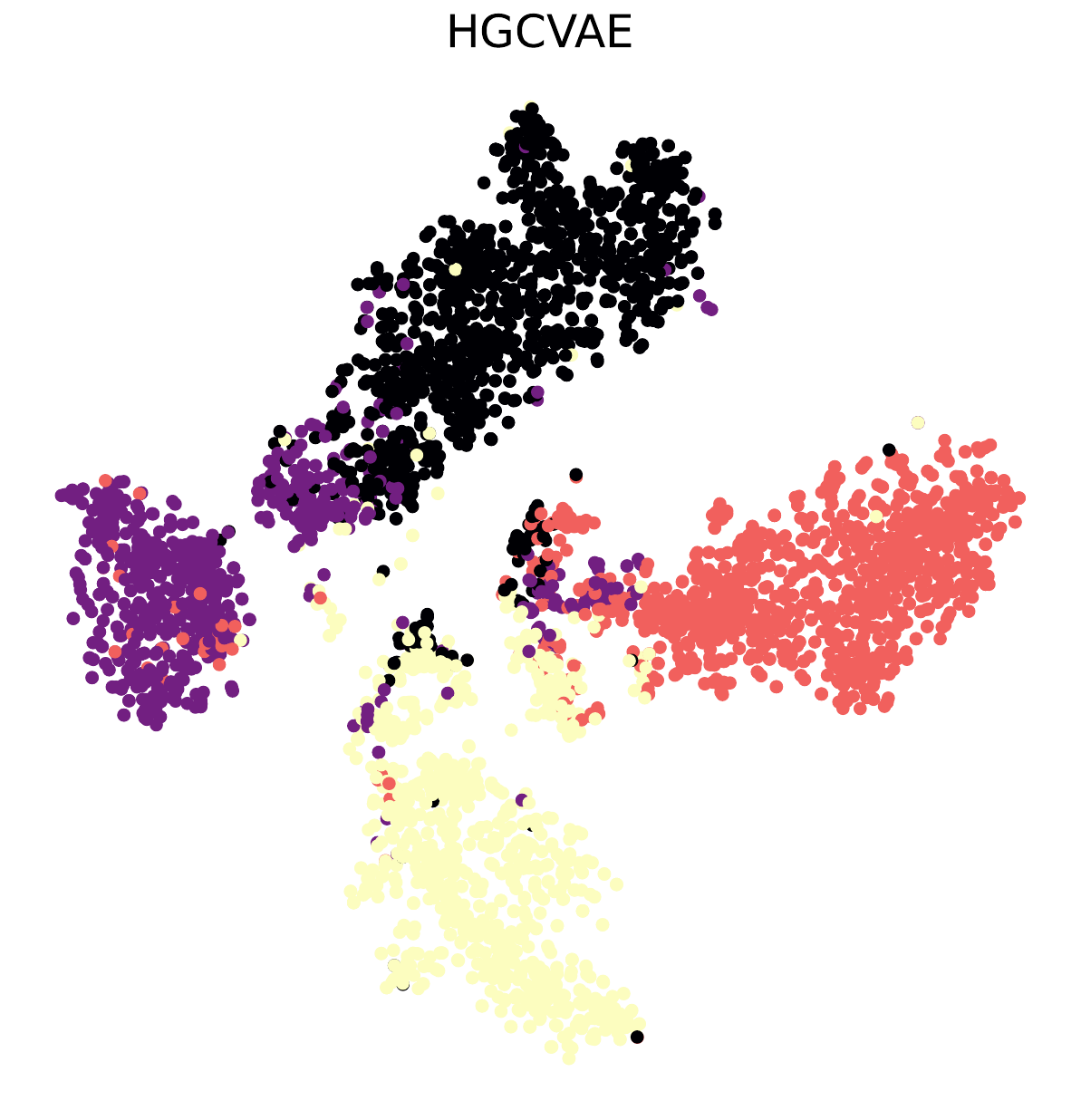}}
  \end{minipage}
  \caption{The embedding visualization of five models and \model. }
  \label{fig:visua}
\end{figure}

To intuitively compare the representations of different baselines, we visualize the node representations on DBLP using t-SNE, which consists of four categories. Figure~\ref{fig:visua} presents the results. We observe that the representations obtained by DGI~\cite{DGI2018}, HeCo~\cite{HeCo}, HGMAE~\cite{HGMAE}, and our model can relatively clearly distinguish the four categories, indicating their effectiveness in capturing discriminative features, while the representations generated by Mp2vec~\cite{mp2vec} and GraphMAE~\cite{GraphMAE} fail to effectively separate the categories. Interestingly, HGMAE categorizes the embeddings into four clusters, but the points within each cluster are dispersed. In contrast, \model efficiently discriminates the four classes with clear boundaries and forms more compact clusters., which again provide evidence of the efficiency and effectiveness of \model.

\section{CONCLUSION AND DISCUSSION}
In this paper, we propose \model to address the challenges in HGL. Leveraging the potential of generative SSL, we concentrate on refining latent representations to enhance graph representation. We introduce a progressive negative sample generation mechanism, which involves utilizes variational inference to generate hard negative samples with shifted stochastic variables. \model demonstrates remarkable performance across various datasets, and thorough ablation studies and parameter analyses further validate the effectiveness of its diverse components.



\bibliographystyle{ieeetr}
\bibliography{ref}

\appendix
\title{Refining Latent Representations: A Generative SSL Approach for Heterogeneous Graph Learning}

\section{Methodology}
\subsection{Notations}
\begin{table}[h]
\centering
\caption{The full notations used in this work.}
\begin{tabular}{c|l}
\toprule
Notation                             & Description                                                                                                          \\ \hline
$\mathcal{V}, \mathcal{E}$           & The node and edge set.                                                                                                    \\
$\phi(\cdot),\psi(\cdot)$            & The mapping function of node and edge types.                                                                          \\
$\mathcal{X},x_v$                    & Node attributes matrix and node feature vector.                                                                       \\
$\gG$                                & A heterogeneous graph.                                                                                                \\
$\gT_{v},\gT_{e}$                    & The sets of possible node and edge types.                                                                             \\
$\rho$                               & A meta-path.                                                                                                          \\
$\rh_{i}$                            &The node representation of node $i$                                                                                      \\
$\mathcal{N}_{i}^{\rho}$             & The neighbors of node $i$ in meta-path $\rho$.                                                                         \\
$att_{\text{node}}$ & The GNN layer for node representation. \\
$att_{\text{sem}}$ & The GNN layer for semantic representation.\\
$e_{ij}^{\rho},\beta_{\rho}$         & \begin{tabular}[c]{@{}l@{}}The importance score between node-pair $(i,j)$ \\ and meta-path.\end{tabular}                   \\
\multirow{2}{*}{$p(h|z)$}            & \multirow{2}{*}{\begin{tabular}[c]{@{}l@{}}The likelihood between embedding $h$ and \\ latent variable $z$.\end{tabular}} \\
                                     &                                                                                                                      \\
$p(z),p(z|h)$                        & The prior and posterior probability respectively.                                                                         \\
$q_{w}(z|h)$                         & The approximation function parameterized by $w$.                                                                         \\
$U,U^{+},U^{-}$                      & The anchor, positive and negative samples.
\\
$\mu,\sigma$                          & The mean and variance.
\\
$\kappa$                              & The shift parameter of $\mu$.
\\
$\mu^{*}$                              & The shifted mean value.
\\
$\lambda$                       & The balance ratio of two kinds of negative samples. \\
$\tau$                          & The temperature parameter for InfoNCE loss.\\
$\gamma$                           & The input feature mask ratio.                            \\ \bottomrule
\end{tabular}
\label{method:notation}
\end{table}

\subsection{Background}~\label{apx:background}
\vpara{Variational graph autoencoder.} Variational graph auto-encoder (VGAE)~\cite{VGAE} is a framework for unsupervised learning on graph-structured data based on the variational auto-encoder (VAE)~\cite{VAE}. VGAE consists variational encoder $f_E$ to map the input heterogeneous graph into dense representation, along with a decoder $f_D$ to recoverr the graph from it. The vaillia VGAE is designed for link-prediction, it takes two-layer GCN as encoder and inner-product as decoder to recover the adjacency matrix, which relies on the the latent variable and is capable of learning interpretable latent representations.

\vpara{Heterogeneous graph attention network}\label{han:encoder} Meta-paths are employed to capture explicit semantic information in the graph. A meta-path $\rho$ is denoted as $\rho=v_0v_1...v_l$, where $v_i$ belongs to different node types $\phi_i$. For example, in the ACM dataset, we design a meta-path P-A-P, indicating that the meta-path starts and ends with a paper node, and is connected by an author node. 

Heterogeneous Graph Attention Network (HAN)~\cite{HAN} is a representative hierarchical HGL model built upon meta-paths. It combines node-level attention for neighbor aggregation with semantic-level attention for semantic aggregation. 

For node-level attention, assume a node-pair $(i,j)$ appearing in a meta-path $\rho$, the importance score between $(i,j)$ is formulated as:
\begin{align}
e_{i j}^{\rho} & = att_{\text{node}}\left(h_i, h_j ; \rho\right),
\end{align}
where $\mathbf{h}_i$ and $\mathbf{h}_j$ represent the node representations of nodes $i$ and $j$, respectively, and $att_{\text{node}}$ denotes the graph attention network (GAT)~\cite{gat} layer responsible for node-level attention. After obtaining the importance scores, the embedding of node $i$ is calculated as:
\begin{equation}
\mathbf{h}_i^{\rho}=\sigma\left(\sum_{j \in \mathcal{N}_i^{\rho}} \text{softmax}_j(e_{i j}^{\rho})  \cdot \mathbf{h}_j\right),
\end{equation}
where $\mathcal{N}_i^{\rho}$ represents the meta-path-based neighbors of node $i$, $h_i^{\rho}$ is the learned embedding of node $i$, and $\sigma$ denotes the activation function. The embeddings of each meta-path are denoted as $\mathcal{H}_{\rho_i}$. 

At the semantic-level aggregation step, we employ another GAT layer, $att_{\text{sem}}$, to calculate the importance score of each meta-path:
\begin{equation}
\left(\beta_{\rho_1}, \ldots, \beta_{\rho_P}\right)= att_{\text{sem}}\left(\mathcal{H}_{\rho_1}, \ldots, \mathcal{H}_{\rho_P}\right).
\end{equation}

Finally, the final embedding $\mathcal{H}$ is obtained in a weighted manner as follows:
\begin{equation}
\mathcal{H}=\sum_{p=1}^P \beta_{\rho_p} \cdot \mathcal{H}_{\rho_p}.
\label{method:han_encode}
\end{equation}

\begin{table}[htp]
\caption{The meta-paths defined for each network.}
\begin{tabular}{@{}ccc@{}}
\hline Heterogeneous graphs & Node types  & Meta-paths                                                  \\ \midrule
ACM                  & \begin{tabular}[c]{@{}c@{}}P: Paper \\ A:Author\\ S:Subject\end{tabular}                                                                  & \begin{tabular}[c]{@{}c@{}}P-A-P\\ P-S-P\end{tabular}       \\ \midrule
DBLP                 & \begin{tabular}[c]{@{}c@{}}P: Paper\\ A: Author\\ T: Term\\ V:  Venue\end{tabular}                                                        & \begin{tabular}[c]{@{}c@{}}APA\\ APVPA\\ APTPA\end{tabular} \\ \midrule
Aminer               & \begin{tabular}[c]{@{}c@{}}P: Paper\\ A: Author\\ R: Relationship\end{tabular}                                                            & \begin{tabular}[c]{@{}c@{}}PAP\\ PRP\end{tabular}           \\ \midrule
Freebase             & \begin{tabular}[c]{@{}c@{}}B: Book\\ F: Film\\ L: location\\ M: Music\\ P: Person\\ S: Sport\\ O: Organization\\ U: business\end{tabular} & \begin{tabular}[c]{@{}c@{}}MDM\\ MAM\\ MWM\end{tabular}     \\ \bottomrule
\end{tabular}
\label{method_metapath}
\end{table}

\subsection{Algorithm}\label{algor:training}
\begin{algorithm}[h]
\caption{\model training algorithm.}\label{alg:procedure}
\KwIn{Predefined meta-paths $\rho={\rho_1,  \rho_2...\rho_n}$; Total training epochs $C$.} 
\For{$epoch$ in $C$}{
        {Mask the original feature with a dynamic mask rate $\gamma$}; \\
        {Encode the masked graph twice to obtain the anchor and postive sample;} \\
        {Adopt two $\operatorname{HAN}$ layers to learn the latent variance, calculating the ELBO using~(\ref{elbo})}; \\        
        {Generate the negative samples $U_{D}^{-}$ by randomly drop a fixed ratio of feature of  $\mathcal{H^{+}}$}; \\         
        {Generate $U_{V}^{-}$ through VI with a shifted mean value $\mu^{*}$ using~(\ref{nega:2})}; \\        
        {Calculate the contrastive loss with $\mathcal{H}, \mathcal{H^{+}}, \mathcal{H^{-}}$ using~(\ref{info})}; \\            
        {Calculate the reconstruct ESCE loss using~(\ref{esce})}; \\
        {Applying stochastic gradient descent to minimize the overall loss $\mathcal{L}$}; 
    }  
\end{algorithm}

\subsection{Negative samples generation with shifted stochastic variables.}
\begin{figure}[h]
\centering 
\includegraphics[width=0.3\textwidth]{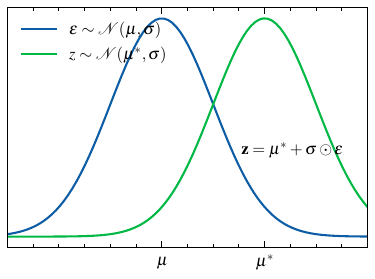} 
\caption{An elaboration of the negative samples generated in $U_{V}^{-}$.} 
\label{fig:vi}
\end{figure}
Figure~\ref{fig:vi} illustrates a comparison between the learned stochastic variables and the shifted stochastic variables, where the latter are utilized for generating hard negative samples. While the variance remains unchanged, the mean value $\mu^{*}$ is slightly shifted using a small coefficient. This adjustment ensures that the latent variables generated by $\mu^{*}$ and $\epsilon$ are similar but belong to distinct categories.


\subsection{Comparison with Previous Works}~\label{method_comparison}
We select three most related works that primarily focus on GAEs to highlight the distinctions between our model and previous studies. These three models are VGAE~\cite{VGAE}, GraphMAE~\cite{GraphMAE}, and HGMAE~\cite{HGMAE}.


\vpara{VGAE.} VGAE represents an early attempt at applying autoencoders to graph-related tasks, specifically for homogeneous graphs in the context of link prediction. VGAE adopts a two-layer GCN as the encoder and uses inner product as the decoder. Our model differs significantly from VGAE in several aspects. Firstly, VGAE is designed for homogeneous graphs, it is challenging to directly apply it to heterogeneous graphs. Secondly, the optimization objectives are different. While VGAE optimizes based on the ELBO and reconstruction constraints, our model employs three strategies: the ELBO constraint, the ESCE constraint, and the contrastive constraint. Thirdly, the ELBO constraint in VGAE and our model is different. VGAE aims to recover the adjacency matrix and focuses on minimizing $\mathrm{D}_{\mathcal{KL}}(q_{\omega}(\mathcal{A} \mid \mathcal{H}) \| p_\theta (\mathcal{A}\mid\mathcal{H} ))$, whereas our model focuses on minimizing $\mathrm{D}_{\mathcal{KL}}(q_{\omega}(\mathcal{Z} \mid \mathcal{H}) \| p_\theta (\mathcal{Z}\mid\mathcal{H} ))$. Fourthly, our model considers a more comprehensive set of graph features and incorporates a series of innovative strategies, such as a dynamic mask strategy and the use of the model's generative capabilities to create high-quality negative samples, thereby enabling more effective learning.

\vpara{GraphMAE.} GraphMAE introduces a masked graph autoencoder that utilizes two rounds of mask operations to ensure robust training and incorporates the SCE as the reconstruction criterion. The core idea of GraphMAE is similar to that of MAE~\cite{he2022masked}, where both methods involve randomly corrupting a portion of the input data and encouraging the model to predict the masked features based on the remaining graph information, such as adjacency knowledge. This marks an early application of MAE in the graph domain.

Our model differs from GraphMAE in several key aspects. Firstly, like VGAE, GraphMAE is designed for homogeneous graphs. Secondly, while GraphMAE employs an autoencoder as its backbone model, our model utilizes a variational autoencoder, harnessing the capabilities of VI for high-quality negative samples generation. Thirdly, the training strategy diverges between the two. While GraphMAE employs a two-round masking strategy to promote robust self-supervised learning and utilizes SCE as the reconstruction loss, our experiments have shown that the re-masking strategy does not contribute to overall performance. In contrast, our training strategy encompasses various novel components, including generative contrastive learning strategies and an enhanced version of the SCE loss, among others.

\vpara{HGMAE.} HGMAE can be credited as a pioneer in utilizing generative models for heterogeneous learning. It introduces meta-path and attribute masking to facilitate stable training and employs three training strategies, including meta-path-based edge reconstruction, target attribute restoration, and positional feature prediction.

However, we harbor reservations regarding the three training strategies employed in HGMAE, particularly concerning the prediction of positional features. HGMAE initially utilizes Mp2vec~\cite{mp2vec} to characterize positional features and obtain $P$. Subsequently, it decodes the model using an MLP decoder to obtain $P1$, and finally, the difference between $P$ and $P1$ is learned through an SCE-like loss. This approach raises potential issues. Firstly, the positional information by Mp2vec is rather limited, particularly in cases where there are only a finite number of meta-paths in each dataset. As a result, the benefits of predicting positional information based on features obtained from Mp2vec encoding are questionable. Secondly, HGMAE proposes to replace the original HAN decoder with an MLP, aiming to exclude positional information during decoding. However, we doubt that HAN itself utilizes GAT as the backbone model to compute attention scores at both the node and semantic levels, the entire computation process does not involve saving or computing positional information. 

In contrast to HGMAE, our approach focuses on addressing core problems in heterogeneous graph learning through more concise training strategies. We retain the feature masking mechanism for robustness and leverage contrastive learning to enhance model performance. We innovatively develop \textbf{PNSM} to produce high-quality negative samples, facilitating effective model learning.

\begin{table*}[t]
\caption{Ablation Studies of the three loss involved in training on model effectiveness.}
\label{apx:ablation_strategies}
\begin{tabular}{@{}cccccccccc@{}}
\toprule
Split               & Datasets                         & \multicolumn{2}{c}{DBLP}                    & \multicolumn{2}{c}{Freebase}                & \multicolumn{2}{c}{ACM}                     & \multicolumn{2}{c}{Aminer}                  \\ \midrule
                    & Metrics                          & Mi-F1                & Ma-F1                & Mi-F1                & Ma-F1                & Mi-F1                & Ma-F1                & Mi-F1                & Ma-F1                \\ \midrule
\multirow{4}{*}{20} & w.o. $\gL_{pnsm}$                & 89.16± 0.85          & 88.46± 0.12          & 60.84± 0.33          & 50.86± 0.31          & 73.08± 2.52          & 66.37± 5.04          & 44.80± 0.41          & 42.27± 0.35          \\
                    & w.o. $\gL_{elbo}$ & 92.30± 0.21          & 91.86± 0.21          & 60.74± 1.14          & 54.73± 0.33          & 87.08± 2.28          & 87.52± 1.84          & 80.38± 0.17          & 72.38± 0.23          \\
                    & w.o. $\gL_{esce}$                & 91.88± 0.12          & 92.20± 0.00          & 58.42± 1.33          & 54.81± 0.68          & 85.70± 1.31          & 86.18± 0.81          & 71.64± 0.37          & 62.53± 0.45          \\
                    & \textbf{\model}                  & \textbf{92.62± 0.18} & \textbf{92.98± 0.12} & \textbf{68.90± 0.81} & \textbf{64.18± 0.84} & \textbf{88.62± 0.26} & \textbf{89.25± 0.57} & \textbf{80.98± 0.07} & \textbf{72.86± 0.12} \\ \midrule
\multirow{4}{*}{40} & w.o. $\gL_{pnsm}$                & 88.74± 0.98          & 88.30± 0.94          & 59.58± 1.88          & 50.61± 1.50          & 74.98± 1.75          & 67.98± 5.31          & 51.58± 0.12          & 47.90± 0.11          \\
                    & w.o. $\gL_{elbo}$                & 88.38± 0.04          & 88.98± 0.00          & 57.96± 1.78          & 52.72± 0.38          & 86.78± 1.02          & 90.01± 0.78          & 81.26± 0.08          & 74.17± 0.51          \\
                    & w.o. $\gL_{esce}$                & 89.34± 0.04          & 89.78± 0.04          & 65.90± 0.91          & 61.25± 0.82          & 89.04± 0.52          & 88.38± 0.67          & 73.84± 0.65          & 63.25± 1.06          \\
                    & \textbf{\model}                  & \textbf{91.86± 0.20} & \textbf{91.65± 0.05} & \textbf{66.04± 1.63} & \textbf{61.21± 1.13} & \textbf{90.40± 0.34} & \textbf{90.26± 0.38} & \textbf{81.32± 0.07} & \textbf{74.87± 0.10} \\ \midrule
\multirow{4}{*}{60} & w.o. $\gL_{pnsm}$                & 89.66± 0.90          & 88.54± 0.89          & 60.28± 1.59          & 50.10± 0.60          & 73.62± 3.30          & 67.18± 6.66          & 54.82± 0.12          & 51.53± 0.16          \\
                    & w.o. $\gL_{elbo}$                & 92.00± 0.00          & 91.19± 0.00          & 58.92± 1.55          & 54.71± 0.45          & 91.00± 0.21          & 91.02± 0.17          & 82.14± 0.08          & 75.84± 0.07          \\
                    & w.o. $\gL_{esce}$                & 92.10± 0.00          & 91.29± 0.00          & 62.18± 1.75          & 57.69± 0.71          & 90.44± 0.19          & 90.66± 0.23          & 71.88± 0.17          & 62.42± 0.26          \\
                    & \textbf{\model}                  & \textbf{94.04± 0.05} & \textbf{93.37± 0.06} & \textbf{68.30± 1.33} & \textbf{62.10± 0.40} & \textbf{92.10± 0.00} & \textbf{92.22± 0.00} & \textbf{82.66± 0.15} & \textbf{76.43± 0.10} \\ \bottomrule
\end{tabular}
\end{table*}

\begin{table*}[t]
\caption{Ablation Studies of different methods of negative samples construction.}
\label{apx:abla_negative_construction_all}

\begin{tabular}{@{}cccccccccc@{}}
\toprule
Split               & Datasets                  & \multicolumn{2}{c}{DBLP}                    & \multicolumn{2}{c}{Freebase}                & \multicolumn{2}{c}{ACM}                     & \multicolumn{2}{c}{Aminer}                  \\ \midrule
                    & Metrics                   & Mi-F1                & Ma-F1                & Mi-F1                & Ma-F1                & Mi-F1                & Ma-F1                & Mi-F1                & Ma-F1                \\ \midrule
\multirow{5}{*}{20} & $\operatorname{\model}_0$ & 88.64± 1.05          & 88.68± 1.48          & 47.68± 0.35          & 42.07± 0.49          & 70.42± 1.54          & 60.69± 5.60          & 49.08± 8.55          & 35.07± 2.61          \\
                    & $\operatorname{\model}_1$ & 90.94± 0.05          & 90.55± 0.05          & 57.30± 0.50          & 50.88± 0.42          & 66.82± 1.53          & 67.78± 1.01          & 70.88± 0.53          & 61.71± 0.75          \\
                    & $\operatorname{\model}_2$ & 87.62± 2.53          & 84.81± 4.40          & 45.20± 0.46          & 40.81± 0.47          & 76.92± 0.19          & 72.76± 0.30          & 55.68± 0.27          & 36.17± 4.96          \\
                    & $\operatorname{\model}_3$ & 92.18± 0.16          & 91.79± 0.16          & 59.96± 1.18          & 47.51± 1.87          & 69.84± 1.43          & 71.24± 0.03          & 70.26± 0.53          & 60.85± 0.86          \\
                    & \textbf{\model}           & \textbf{92.62± 0.18} & \textbf{92.98± 0.12} & \textbf{68.90± 0.81} & \textbf{64.18± 0.84} & \textbf{88.62± 0.26} & \textbf{89.25± 0.57} & \textbf{80.98± 0.07} & \textbf{72.86± 0.12} \\ \midrule
\multirow{5}{*}{40} & $\operatorname{\model}_0$ & 88.34± 0.62          & 87.45± 0.89          & 49.92± 2.52          & 40.91± 0.93          & 69.74± 2.87          & 59.81± 2.39          & 52.68± 0.24          & 39.54± 6.97          \\
                    & $\operatorname{\model}_1$ & 89.02± 0.04          & 88.75± 0.04          & 54.90± 0.73          & 51.32± 0.39          & 80.26± 0.15          & 78.86± 0.18          & 73.12± 0.70          & 62.78± 1.10          \\
                    & $\operatorname{\model}_2$ & 87.62± 0.21          & 86.48± 0.33          & 47.32± 1.53          & 39.30± 0.91          & 76.36± 0.55          & 70.62± 0.51          & 53.76± 2.73          & 40.69± 6.53          \\
                    & $\operatorname{\model}_3$ & 89.62± 0.04          & 89.41± 0.14          & 51.94± 1.60          & 45.79± 0.47          & 83.06± 0.10          & 82.01± 0.12          & 71.90± 0.28          & 61.64± 0.24          \\
                    & \textbf{\model}           & \textbf{91.86± 0.20} & \textbf{91.65± 0.05} & \textbf{68.90± 0.81} & \textbf{64.18± 0.84} & \textbf{88.62± 0.26} & \textbf{89.25± 0.57} & \textbf{80.98± 0.07} & \textbf{72.86± 0.12} \\ \midrule
\multirow{5}{*}{60} & $\operatorname{\model}_0$ & 89.46± 0.42          & 87.30± 0.70          & 49.38± 3.50          & 41.20± 2.26          & 72.32± 1.79          & 68.53± 1.77          & 55.30± 4.99          & 47.56± 3.20          \\
                    & $\operatorname{\model}_1$ & 90.70± 0.09          & 90.02± 0.09          & 58.22± 0.45          & 47.79± 0.82          & 80.68± 0.10          & 80.37± 0.07          & 71.28± 0.71          & 61.94± 0.61          \\
                    & $\operatorname{\model}_2$ & 89.46± 0.05          & 87.39± 0.11          & 48.12± 1.17          & 43.19± 0.85          & 74.46± 2.29          & 69.25± 4.46          & 54.86± 1.86          & 48.67± 1.65          \\
                    & $\operatorname{\model}_3$ & 91.42± 0.04          & 90.66± 0.05          & 54.40± 2.22          & 46.33± 0.56          & 83.44± 0.08          & 83.68± 0.08          & 70.80± 0.19          & 61.75± 0.60          \\
                    & \textbf{\model}           & \textbf{94.04± 0.05} & \textbf{93.37± 0.06} & \textbf{68.90± 0.81} & \textbf{64.18± 0.84} & \textbf{88.62± 0.26} & \textbf{89.25± 0.57} & \textbf{80.98± 0.07} & \textbf{72.86± 0.12} \\ \bottomrule
\end{tabular}
\end{table*}

\section{Experiments}
 \begin{table}[hb]
\centering
\setlength{\tabcolsep}{3.5mm}
\caption{Statistics of datasets used in this paper.}
\resizebox{\linewidth}{!}{ 
\begin{tabular}{@{}ccccc@{}}
\hline Dataset  & Nodes  & Edges   & Type             & Target \\ 
\hline DBLP     & 26,128 & 39,290 &Citation Network
 & Author \\
ACM      & 11,246 & 34,852  & Citation Network  & Paper  \\
Freebase & 43,854 & 151,043  &  Knowledge Graph  & Movie  \\
AMiner   & 55,783 & 153,676  & Citation Network   & Paper  \\ 
\hline
\end{tabular}}
\vspace{5pt}
\label{data:info}
\end{table}

\subsection{Experiment Settings}~\label{apx:experiment_settings}

\vpara{Datasets.}
We employ four real-world datasets inculding DBLP~\cite{magnn}, ACM~\cite{acm}, Freebase~\cite{freebasedata} and Ainer~\cite{aminer_herec} to evaluate the model performance. The detailed statistics of these datasets are summarized in Table~\ref{data:info}.

\vpara{Baselines.}
To make comprehensive comparison, we compare \model with eleven baseline models of four types, including two network embedding models, GraphSAGE~\cite{graphsage} and Mp2vec~\cite{mp2vec}, two contrative-oriented models, DGI~\cite{DGI2018} and DMGI~\cite{DMGI}, four heterogeneous graph neural networks, HERec~\cite{aminer_herec}, HetGNN~\cite{hetgnn}, HAN~\cite{HAN} and HeCo~\cite{HeCo}, three graph autoencoder based models, VGAE~\cite{VGAE}, GraphMAE~\cite{GraphMAE} and HGMAE~\cite{HGMAE}. 

\vpara{Experimental setups.} 
For \model, the main settings are consistent with the prior settings~\cite{HeCo,GraphMAE,HGMAE}. Please noted that we report the performance of all baselines under best parameter settings provided in their original papers.

We list the main parameters of \model in Table~\ref{param:info}. We use HAN~\cite{HAN} as the default encoder and decoder, the learning rate is searched between 2e-4 and 2e-3, the deviation value $\kappa$ is 2 and the power factor of $\delta$ is 3. For node classification task, we use Micro-F1 and Macro-F1 as the evaluation metric, and we report normalized mutual information (NMI) and adjusted rand index (ARI) for node clustering task. All experiments conduct on a single A100 80G and the results are reported by repeating the experiments five times for each method to take the average and deviation.

\begin{table}[htb]
\centering
\setlength{\tabcolsep}{2.5mm}
\caption{The settings of main parameter.}
\begin{tabular}{c|cccc}
\hline Params & E/Decoder & LR             & $\kappa$ & $\delta$  \\
\hline Value  & HAN       & 2e-4$\sim$2e-3          & 2        & 3       \\ \hline
\end{tabular}
\vspace{5pt}

\label{param:info}
\end{table}
\subsection{Ablation Study}~\label{apx:ablation_studies}
The ablation studies of training losses and different negative samples construction methods are presented in this subsection.
\subsubsection{Ablation Studies of training loss.}~\label{apx:ablation_studies_training_losses}
We present a series of experiments about the training losses impact, and the main results can be found in Table~\ref{apx:ablation_strategies}, which further validates the effectiveness of our approach.

\subsubsection{Ablation Studies of different negative samples construction methods.}~\label{apx:ablation_studies_negative_construction}
We present a series of experiments about the different negative samples construction methods. And our main results show in Table~\ref{apx:abla_negative_construction_all}.
\subsection{Parameter analysis}~\label{apx:para_analysis}
The analysis of the impact of hidden dimension, mean shift factor and the number of negative samples are presented in this subsection.
\begin{figure*}[ht]
\includegraphics[width=0.47\columnwidth]{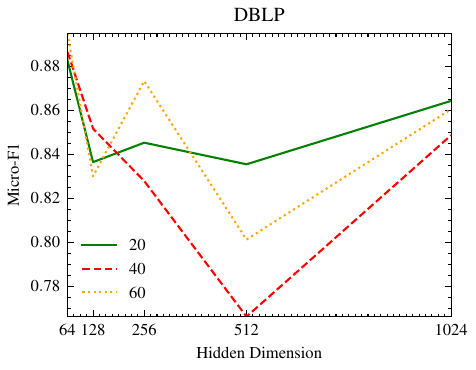}
\includegraphics[width=0.47\columnwidth]{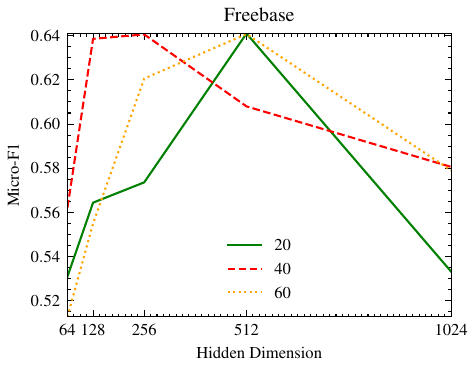}
\includegraphics[width=0.47\columnwidth]{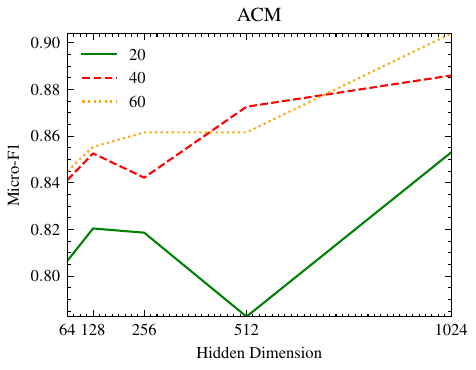}
\includegraphics[width=0.47\columnwidth]{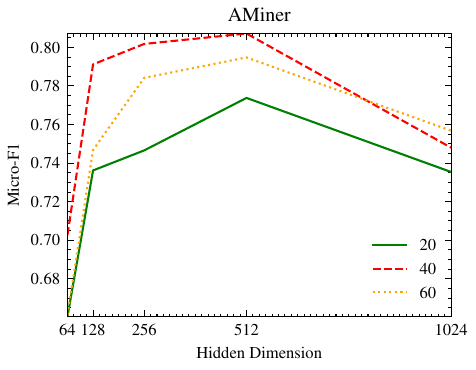}
  \caption{Performance of \model w.r.t. hidden dimension.}
  \label{apx:fig_hidden}
\end{figure*}

\begin{figure*}[ht]
\includegraphics[width=0.47\columnwidth]{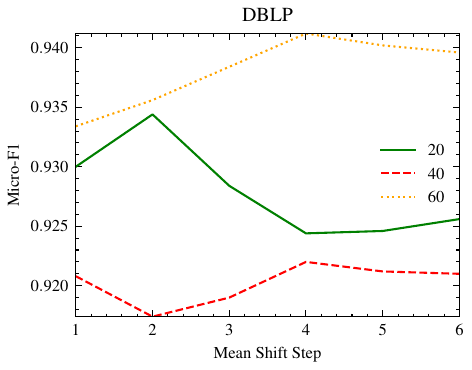}
\includegraphics[width=0.47\columnwidth]{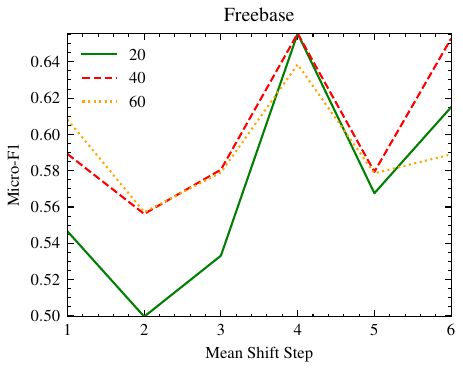}
\includegraphics[width=0.47\columnwidth]{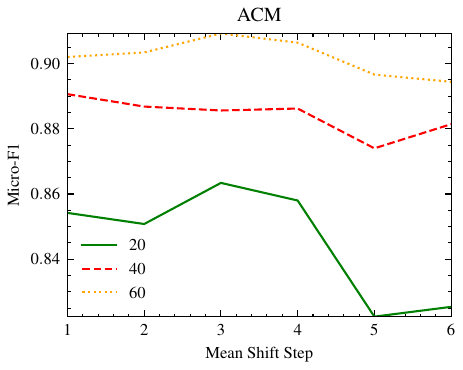}
\includegraphics[width=0.47\columnwidth]{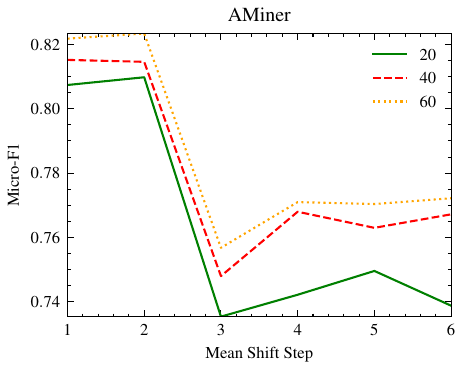}
 
  \caption{Performance of \model w.r.t. mean shift factor. }
  \label{apx:fig_shift}
\end{figure*}

\begin{figure}[!htbp]
\includegraphics[width=0.45\columnwidth]{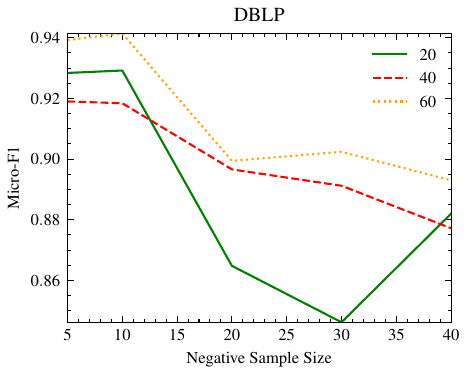}
\includegraphics[width=0.45\columnwidth]{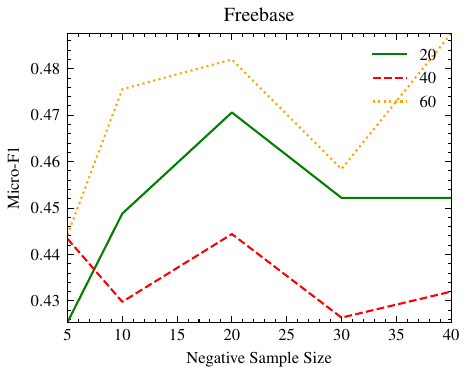}
\includegraphics[width=0.45\columnwidth]{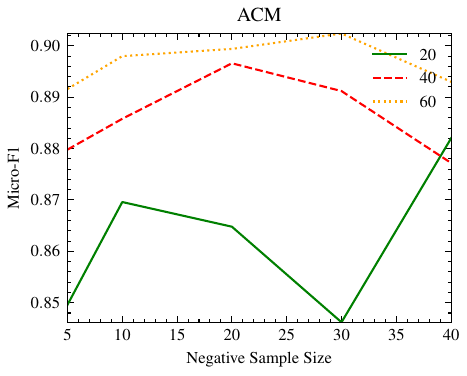}
\includegraphics[width=0.45\columnwidth]{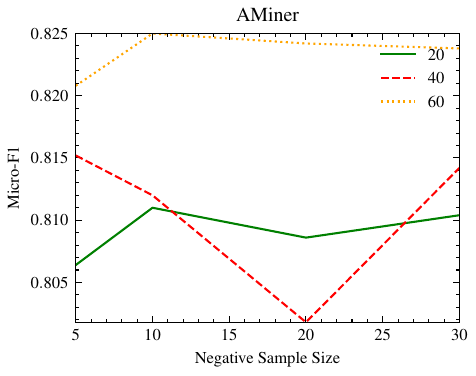}
 
  \caption{Performance of \model w.r.t. negative samples size. }
  \label{apx:fig_neg}
\end{figure}
\subsubsection{Impact of hidden dimension.}~\label{apx:para_analysis_hidden_dim}
We further validate the ablation experiments about model hidden dimension, which shows in Figure~\ref{apx:fig_hidden}.

\subsubsection{Impact of mean shift factor.}~\label{apx:para_analysis_mean_shift}
Next, we show the results about the impact of mean shift factor in Figure~\ref{apx:fig_shift}

\subsubsection{Impact of the number of negative samples.}~\label{apx:para_analysis_negative_samples}
Finally, the impact of the number of negative samples results present in Figure~\ref{apx:fig_neg}

\clearpage
\end{document}

%% file: math_commands.tex
\usepackage{amsmath,amsfonts,bm}

\def\eqref#1{equation~\ref{#1}}

\def\1{\bm{1}}

\def\rh{{\textnormal{h}}}

\DeclareMathAlphabet{\mathsfit}{\encodingdefault}{\sfdefault}{m}{sl}
\SetMathAlphabet{\mathsfit}{bold}{\encodingdefault}{\sfdefault}{bx}{n}

\def\gG{{\mathcal{G}}}
\def\gH{{\mathcal{H}}}

\def\gL{{\mathcal{L}}}

\def\gT{{\mathcal{T}}}

